**Improving the accuracy and generalizability of molecular property regression models with a substructure-substitution-rule-informed framework**

Xiaoyu Fan[1], Lin Guo[1], Ruizhen Jia[1], Yang Tian[1], Zhihao Yang[1], Weihao Li[2] and Boxue Tian[1#]

*[1]MOE Key Laboratory of Bioinformatics, State Key Laboratory of Molecular Oncology, Beijing Frontier Research Center for Biological Structure, School of Pharmaceutical Sciences, Tsinghua University, Beijing, 100084, China. [2]Department of Statistics and Data Science, Tsinghua University, Beijing, 100084, China.*

[#] To whom correspondence should be addressed:

Boxue Tian: boxuetian@mail.tsinghua.edu.cn

**Abstract**

Artificial Intelligence (AI)-aided drug discovery is an active research field, yet AI models often exhibit poor accuracy in regression tasks for molecular property prediction, and perform catastrophically poorly for out-of-distribution (OOD) molecules. Here, we present MolRuleLoss, a substructure-substitution-rule-informed framework that improves the accuracy and generalizability of multiple molecular property regression models (MPRMs) such as GEM and UniMol for diverse molecular property prediction tasks. MolRuleLoss incorporates partial derivative constraints for substructure substitution rules (SSRs) into an MPRM's loss function. When using GEM models for predicting lipophilicity, water solubility, and solvation-free energy (using lipophilicity, ESOL, and freeSolv datasets from MoleculeNet), the root mean squared error (RMSE) values with and without MolRuleLoss were 0.587 vs. 0.660, 0.777 vs. 0.798, and 1.252 vs. 1.877, respectively, representing 2.6–33.3% performance improvements. We show that both the number and the quality of SSRs contribute to the magnitude of prediction accuracy gains obtained upon adding MolRuleLoss to an MPRM. MolRuleLoss improved the generalizability of MPRMs for "activity cliff" molecules in a lipophilicity prediction task and improved the generalizability of MPRMs for OOD molecules in a melting point prediction task. In a molecular weight prediction task for OOD molecules, MolRuleLoss reduced the RMSE value of a GEM model from 29.507 to 0.007. We also provide a formal demonstration that the upper bound of the variation for property change of SSRs are positively correlated with an MPRM's error. Together, we show that using the MolRuleLoss framework as a bolt-on boosts the prediction accuracy and generalizability of multiple MPRMs, supporting diverse applications in areas like cheminformatics and AI-aided drug discovery.

## Introduction

ADMET (Absorption, Distribution, Metabolism, Excretion, and Toxicity) properties impact a compound's potential for therapeutic application[1, 2]. Compared to experimental approaches, Molecular Property Prediction Models (MPPMs) can in theory reduce costs and increase efficiency for optimizing ADMET properties[3-5]. Trained humans can, based on their experience, visually evaluate the structures of compounds predicted by an MPPM, enabling selection of compounds with desired properties[6-8]. Early MPPMs were based on (for example) empirical formulas, quantum chemistry calculations, and machine learning (ML) approaches[9-12]. Recent advances in deep learning (DL) have supported the development of more sophisticated MPPMs based on graph neural networks (GNNs) and Transformers[13-16]. Notably, the development of these MPPMs can be fully data-driven, and these DL-based MPPMs have achieved state-of-the-art (SOTA) performance on benchmarking datasets (*e.g.*, from MoleculeNet and MolData).

Classification models, a subtype of MPPMs, can predict a compound's status in terms of various categories (*e.g.*, whether a compound is toxic or non-toxic; whether it can pass the blood brain barrier, etc.). Molecular Property Regression Models (MPRMs), which are another subtype of MPPM that often use Simplified Molecular Input Line Entry System (SMILES) string[17], molecular graph[18, 19], and 3D coordinates[20, 21] as input, can predict a continuous numerical value (*e.g.,* the solubility of a compound; the lipophilicity of a compound). While data scarcity is common in ADMET predictions, MPRMs have achieved high accuracy when using pre-training with large-scale datasets of unlabeled compounds followed by fine-tuning using datasets specific to desired target properties. For example, the GEM model uses both bond length and angle information for pretraining and the Uni-Mol model incorporates atomic distance information for pretraining[22, 23].

Given that MPRMs are now integral throughout chemometrics, the ability to improve their prediction performance would be greatly welcomed. However, DL-based MPRMs suffer from overfitting on small datasets[24]. This issue manifests in two scenarios: for in-distribution (ID) data and for out-of-distribution (OOD) data[25-27]. The ID overfitting scenario often relates to structurally similar molecules having large variance in properties, a situation termed "activity cliffs"[28, 29]. Common solutions for addressing activity cliffs include self-supervised pretraining[30, 31], data augmentation[32, 33], and adding domain knowledge[34, 35]. The ODD overfitting scenario can be summarized as poor predictive accuracy of DL-based MPRMs on OOD data vs. ID data[36-38]. This performance gap is evident across diverse DL architectures and for various molecular property prediction tasks[39-41]

Despite the strong promise of MPRMs, in practice it is still common for chemists to harness their practical domain knowledge when optimizing ADMET properties in a given real-world project. A typical human domain knowledge is heuristics that predict a potential property change upon substituting one substructure with another. For example, to enhance solubility, a chemist might substitute a hydrophobic group like a phenyl with a piperazine; that is, by assuming that the $\Delta_{solubility}$ for phenyl→piperazine will predictably increase. Analogous to these heuristics, the concept of substructure substitution rules (SSRs) in chemoinformatics is formalized through defining specific transforms (or reaction rules) that dictate how molecular substructures can be interchanged while maintaining or altering desired properties. We speculated that the prediction performance of MPRMs could potentially be improved by incorporating SSRs.

In this study, we introduce MolRuleLoss, a substructure-substitution-rule-informed framework designed to enhance the accuracy and generalizability of MPRMs by incorporating constraints for SSRs into an MPRM's loss function. MolRuleLoss boosts the accuracy of state-of-the-art (SOTA) MPRMs (GEM and UniMol) for diverse molecular property prediction tasks, including predicting lipophilicity, water solubility and solvation-free energy. We show that both the number and quality of rules contribute to the magnitude of prediction accuracy gains obtained upon adding MolRuleLoss to an MPRM. Moreover, MolRuleLoss improves the generalizability of MPRMs for "activity cliff" and OOD molecules. In an extreme scenario for molecular weight prediction addressing the OOD problem, MolRuleLoss reduces the root mean squared error (RMSE) of a GEM model from 29.507 to 0.007. Finally, upon considering the prediction performance gain data across the multiple tasks we examined in this study, we discovered an informative and practically useful trend—that model error increases along with increases in the $STD_{max}$ value (*i.e.*, property change variation)—which we formalized as a conjecture with a mathematical proof. Our study shows that adding the MolRuleLoss framework enables more accurate MPRM predictions without changing the original MPRM's architecture, like a bolt-on to support diverse MPRMs for a wide range of molecular property prediction applications in cheminformatics.

## RESULTS

### MolRuleLoss adds substructure substitution rules from the MMPA method to the loss function of MPRMs

Previous studies of molecular property prediction have shown that incorporating knowledge can improve predictive accuracy[8, 42]. In many MPRMs, a standard mean squared error (MSE) term that captures the difference between predicted and true labels is often used as a loss function, providing a signal for optimizing model parameters[43-45]. Inspired by Physics-Informed Neural Networks (PINNs)[46, 47], which embed physical formulas into DL models, we speculated that using partial derivatives to impose constraints based on a set of substructure substitution rules for a specific task (*e.g.*, predicting lipophilicity and solubility) would potentially improve MPRM accuracy. Pursuing this, we designed the MolRuleLoss framework, leveraging a set of SSRs derived from the matched molecular pair analysis (MMPA) method[48] seeking to enhance the accuracy of various MPRMs, ideally for multiple prediction tasks. Our MolRuleLoss framework allows for the automated extraction of SSRs directly from chemical datasets, which can subsequently be applied as constraints to an MPRM.

MolRuleLoss comprises an MSE term and an SSR term for quantifying property changes for the replacement of one molecular substructure with another (Fig. 1a).

Lambda (λ) weighs the SSR term relative to the MSE term in the loss function (Eqn. 1). Notably, our experience ultimately showed that λ should be adjusted according to the specific desired prediction task. The default value of λ in MolRuleLoss is 0.3. When implementing an SSR in the loss function, we relate the output value of a DL model, y, to the change in the number of $x_i$ and $x_j$ substructures (Eqn. 2). The n-D vector saves the number of substructures for each molecule, which is converted into SSRs automatically within the MolRuleLoss framework (Fig. 1a).

$$Loss = Loss_{MSE} + \lambda \times Loss_{SSR} \qquad (Eqn. 1)$$

$$Loss_{SSR} = \sum_{1 \le i,j \le n} \left[ \left( \frac{\partial y}{\partial \# x_i} - \frac{\partial y}{\partial \# x_j} \right) - rule(x_i, x_j) \right]^2 \quad (Eqn. 2)$$

The specific SSRs we implemented in MolRuleLoss are based on pairwise comparison of two molecules in the training dataset (that is, excluded from the test set, seeking to prevent data leaking); these two molecules are identical except for a single substructure substitution. For instance, with a common molecular scaffold "R", one molecule might have "A" attached (R-A) and another "B" (R-B) (Fig. 1b). The structural change from A to B induces a property change ($\Delta P = P_{[R-A]} - P_{[R-B]}$), like solubility or lipophilicity (logP), which yields a quantifiable rule for the substitution of substructure A with substructure B that corresponds to a property change (ΔP).

Considering that the property changes caused by substructure transformations vary across different molecules, we introduce an adaptive unit. For each rule, a learnable parameter vector with the same dimensionality as the molecular features is added. The dot product between the molecular feature vector and this learnable parameter vector serves as the adaptive unit, which is incorporated into the SSR term in the loss function. Mathematically, such SSRs can be modeled using partial derivatives, where the derivative of the molecular property with respect to the count of substructure A, minus the derivative with respect to substructure B, equals ΔP. Statistical insights, such as the average property change and its variability (defined by the STD) can be calculated (Fig. 1a). MolRuleLoss is a framework that can be applied as a bolt-on to diverse MPRMs.

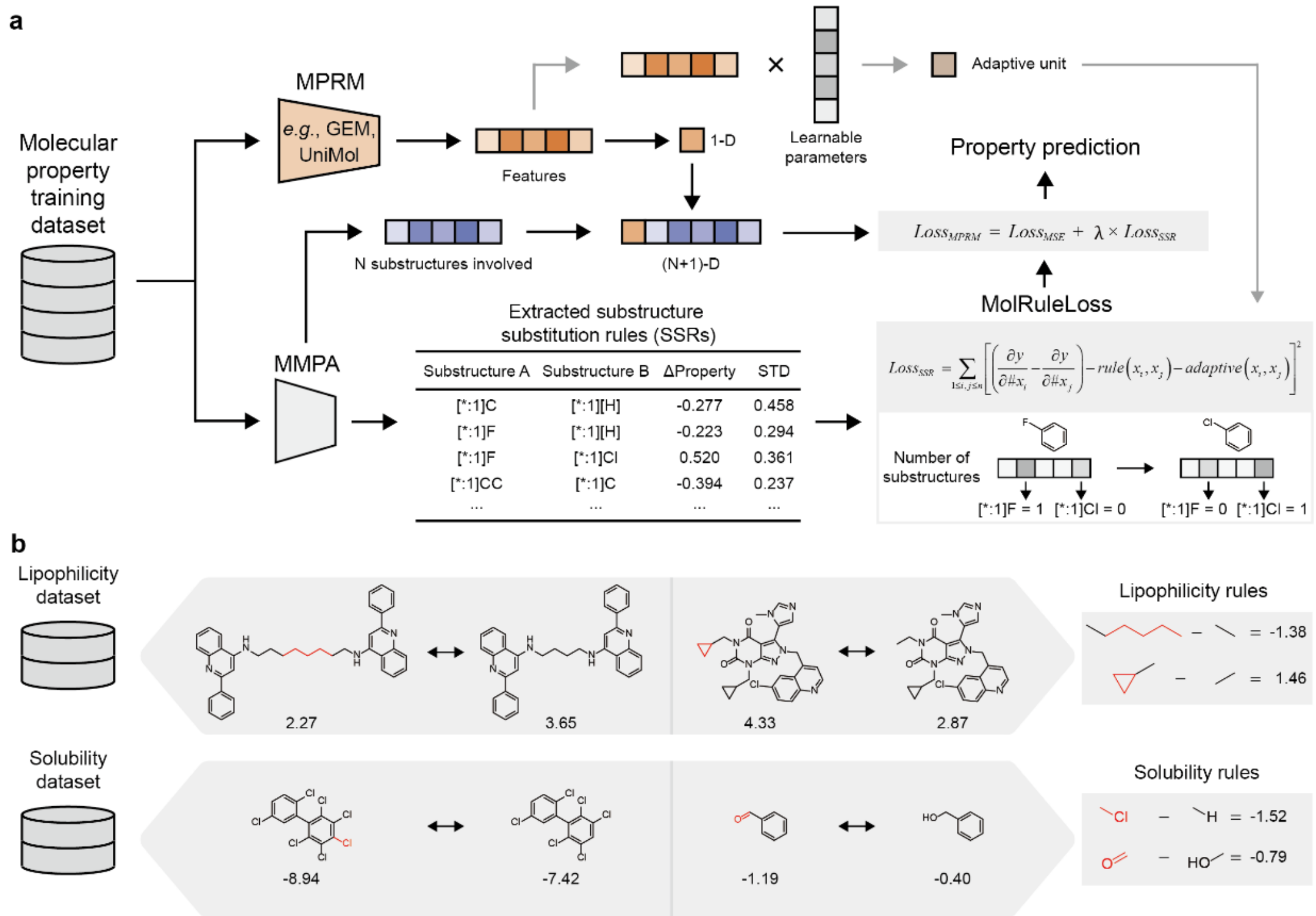

**Figure 1 | Overview of the MolRuleLoss Framework.** Our MolRuleLoss framework leverages a set of substructure substitution rules (SSRs) derived from the matched molecular pair analysis (MMPA) method, seeking to enhance the accuracy of MPRMs for multiple prediction tasks. **a,** MolRuleLoss comprises an MSE term and an SSR term for quantifying the property changes of SSRs. λ controls the relative importance of SSRs for MolRuleLoss's MSE term (Eqn. 1). When implementing an SSR in the loss function of an MPRM, we relate the output value of a DL model, y, to the change in the number of $x_i$ and $x_j$ substructures (Eqn. 2). Extracted SSRs will be converted to rules automatically within the MolRuleLoss framework. **b,** Examples of SSRs. The specific SSRs from the MMPA method are based on pairwise comparison of two molecules in the training dataset; these two molecules are identical except for a single substructure substitution. The structural change from A to B induces a property change ($\Delta P = P_{[R-A]} - P_{[R-B]}$), like solubility or lipophilicity (logP).

## Improving the accuracy of multiple MPRMs for diverse tasks using MolRuleLoss

We initially applied MolRuleLoss to MPRMs for the lipophilicity prediction task, using the MoleculeNet lipophilicity dataset (Lipo), which contains 4,200 entries: each entry is a molecule structure in SMILES format and its corresponding logP value. We divided this dataset into training, validation, and testing sets at an 8:1:1 ratio using the same scaffold split method employed for the development of previously reported MPRMs[49, 50]. We used GEM (Geometry-Enhanced Molecular Representation Learning) as our base model; GEM uses a purpose-built geometry-based graph neural network architecture and a set of dedicated geometry-level self-supervised learning strategies to capture molecular geometry knowledge[18]. Note that the results we report are the average over five random seeds. For the test set, the RMSE of the GEM model with MolRuleLoss was 0.587, compared to 0.660 without it, indicating that MolRuleLoss improves the predictive accuracy of the GEM model (Fig. 2a).

We next trained UniMol and GEM models for predicting lipophilicity, water solubility and solvation-free energy (using the lipophilicity, ESOL, and freeSolv datasets from MoleculeNet), all with and without MolRuleLoss (Fig. 2b-c). For the GEM model on the three tasks, the RMSE values with and without MolRuleLoss were 0.587 vs. 0.660, 0.777 vs. 0.798, and 1.252 vs. 1.877, representing 2.6% to 33.3% performance improvements ($P < 0.001$ for all three tasks, Fig. 2b). Note that Uni-Mol is a pretrained model with the SE(3) Transformer architecture that achieved SOTA performance in multiple MPPM contests[20]. For the three tasks with the UniMol model, the RMSE values with and without MolRuleLoss were 0.568 vs. 0.603, 0.775 vs. 0.788, and 1.395 vs. 1.480 ($P < 0.001$ for all three tasks, 1.9% to 12.6% performance improvements; Fig. 2c). These results show that MolRuleLoss boosts the accuracy of multiple MPRMs for diverse molecular property prediction tasks. It bears emphasis that, given that UniMol and GEM are highly optimized MPPMs with recognized SOTA performance, the increases gained by adding MolRuleLoss would be expected to manifest as substantive improvements for the set of downstream applications (for example in ADMET optimizations) that use the property prediction values as their foundation.

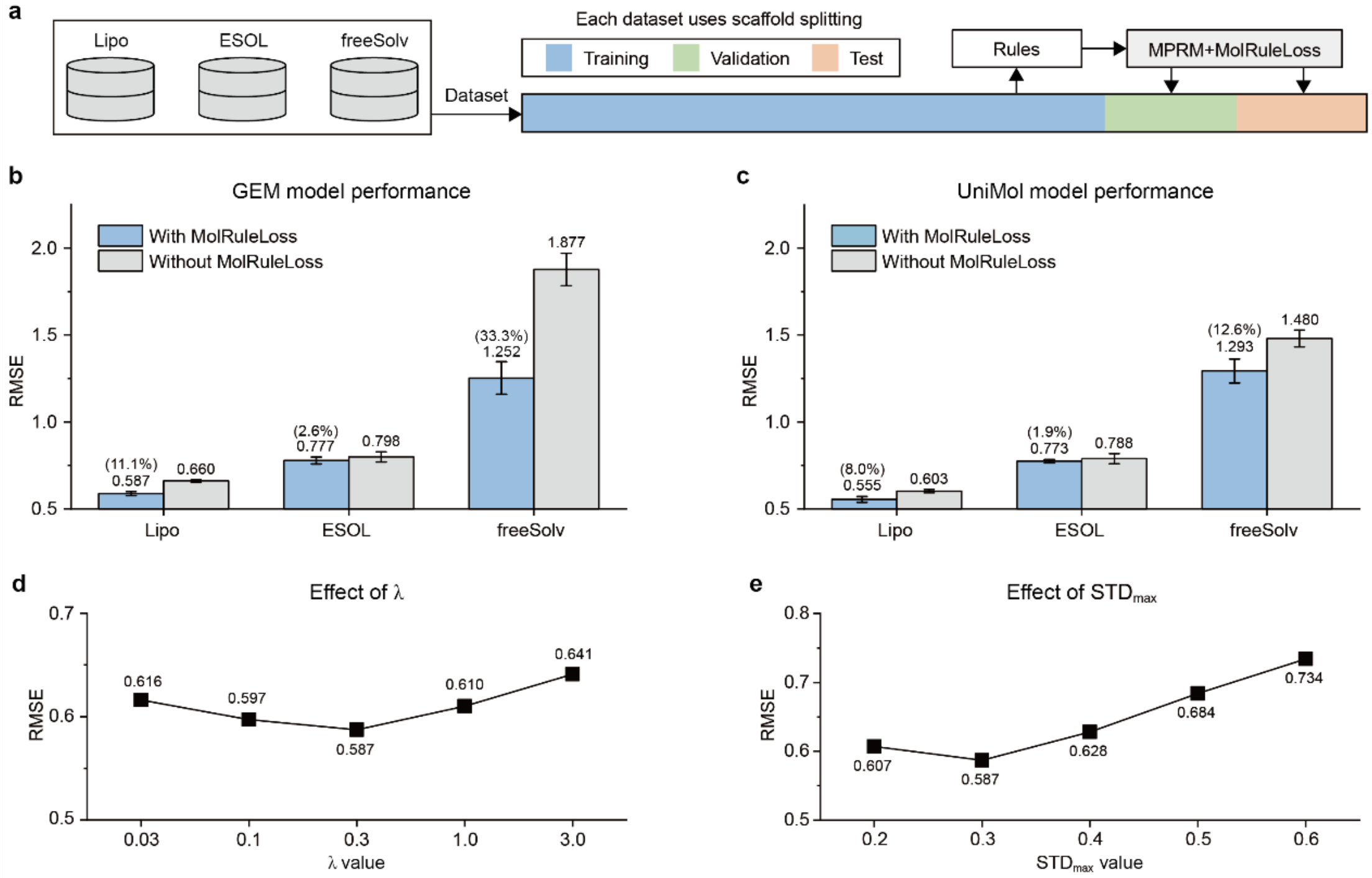

**Figure 2 | Comparison of MPRMs with and without MolRuleLoss. a,** The data splitting scheme and SSR extraction process. **b,** RMSE comparison for GEM models with or without MolRuleLoss, using the lipophilicity, ESOL, and freeSolv datasets from MoleculeNet. **c,** UniMol models were trained for predicting the same tasks, with and without MolRuleLoss. GEM and UniMol are highly optimized MPPMs with state-of-the-art (SOTA) performance on MoleculeNet; our results show that MolRuleLoss boost the accuracy of multiple MPRMs for diverse property prediction tasks. **d-e,** Assessing the influence of MolRuleLoss hyperparameters on lipophilicity prediction performance. **d,** The lipophilicity prediction accuracy of GEM using the indicated distinct λ values; λ weighs the SSR term relative to the MSE term in the loss function. **e,** The lipophilicity prediction accuracy of the GEM models using the indicated $STD_{max}$ values.

## Assessing the influence of MolRuleLoss hyperparameters on prediction accuracy

The accuracy of MPRMs using our MolRuleLoss framework are in theory affected by two hyperparameters: the Lambda value (λ), which weighs the SSR term relative to the MSE term in the loss function (Eqn. 1); and the STD threshold used to decide which SSRs to include ($STD_{max}$), indicating the property variation for the SSR when it alters properties across various molecular scaffolds. We used GEM as the base model for evaluation of hyperparameter influence of MolRuleLoss on the lipophilicity prediction task (using the Lipo dataset). Our analysis started with λ=0.3 and STDmax=0.3, and we only retained SSRs with an occurrence higher than 10 (Methods). We found that λ=0.3 performed best for the lipophilicity prediction task, with an RMSE value of 0.587 on the test set; the RMSEs for λ=3.0 and λ=0.03 were 0.641 and 0.616 (Fig. 2d). These results establish that the accuracy of an MPRM using MolRuleLoss is sensitive to λ.

Given that not all molecules in the testing set possess substructure transformations defined by our SSR set, we expect that the number and quality of rules (defined by $STD_{max}$, with a low value indicating high quality) will affect the accuracy of MPRMs with MolRuleLoss. To this end, we evaluated the influence of $STD_{max}$ (at λ=0.3) with GEM as the base model and using the Lipo dataset. The RMSE values were 0.607 and 0.734 for STDmax=0.2 and STDmax=0.6 (Fig. 2e). When $STD_{max}$ is small, fewer rules are obtained; when $STD_{max}$ is large, low quality rules are obtained. These results show that both the number and quality of rules affect the accuracy of MPRMs.

**The influence of MolRuleLoss training data dimensionality and rule number on prediction accuracy**

Previous studies have established that MPRM accuracy can be improved by increasing dataset volume[40]. In theory, an MPRM can automatically learn "rules" of a task in a "big data" regime (Fig. 3a). To investigate the influence training data dimensionality on the prediction accuracy upon adding MolRuleLoss to an MPRM, we applied MolRuleLoss to a GEM model for predicting the melting point (MP) of compounds. We used an MP dataset comprising 275K entries (each entry comprising a SMILES structure and a temperature value) from OChem[51]. Based on this MP275K dataset, we generated a test set of 175K entries using scaffold splitting, and then extracted SSRs from the training dataset to train MPRMs for accuracy evaluation on the same MP175K test set (Fig. 3b).

We first trained GEM models without MolRuleLoss at different data volumes, including 0.5K, 5K, 50K, and 100K (Fig. 3c). The RMSE values for GEM without MolRuleLoss on the 175K test set were, respectively, 51.372, 46.700, 39.705, and 38.291. These results show that expanding training data dimensionality boosts the accuracy. We then trained GEM models with MolRuleLoss on the 175K test set, and the RMSE values for GEM with MolRuleLoss were 49.082, 43.290, 38.748, and 37.422 ($\lambda$=0.3, $STD_{max}$=0.27) (Fig. 3c). These results show that MolRuleLoss consistently boosts the accuracy of GEM, even upon massively expanding the dimensionality of training data.

Considering that the SSR set derived from a small training dataset is relatively sparse (the number of SSRs for 0.5K is 44, while the number of SSRs for 100K is 150), we speculated that adding more SSRs from another source, *e.g.*, from another dataset for the same property, might improve the prediction accuracy of MPRMs. This is conceptually analogous to "transfer learning"[52]. To test this, we transferred 150 SSRs obtained with MolRuleLoss upon training with the MP100K dataset to an MP0.5K dataset. For the MP0.5K dataset, the GEM model without MolRuleLoss yielded an RMSE of 51.372. When 44 SSRs extracted from MP0.5K were applied through MolRuleLoss, the RMSE improved by 4.4% to 49.082. Using 150 MP100K SSRs as constraints led to an even greater improvement, reducing the RMSE to 46.948, a 8.6% improvement over the GEM without MolRuleLoss (Fig. 3e). These results indicate that expanding the number of rules can improve the accuracy of MPRMs with MolRuleLoss. Notably, while data in drug discovery is generally expensive and time-consuming to obtain, our results show that using rules can apparently help address the data scarcity issue in ADMET optimization efforts.

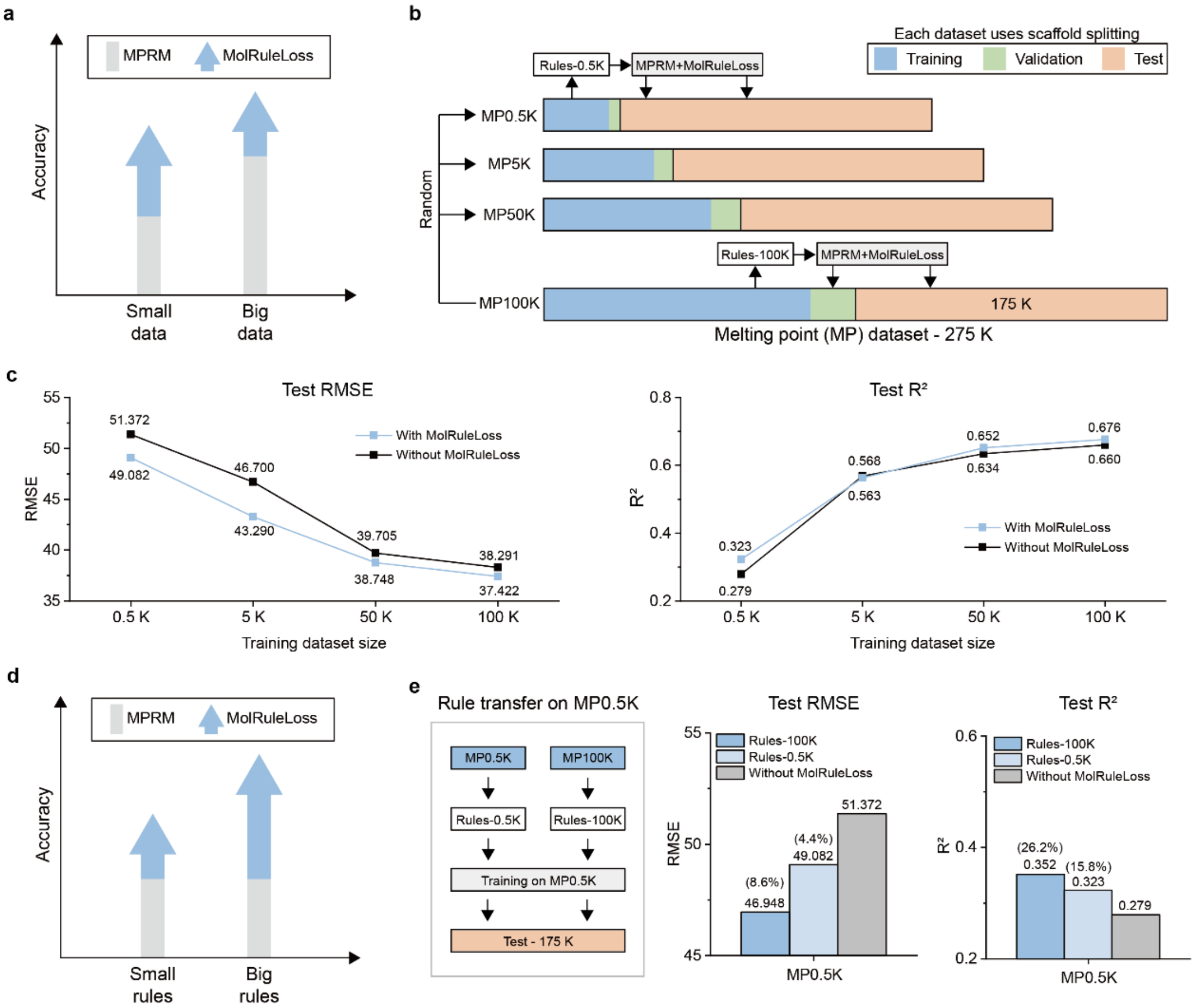

**Figure 3 | Impact of training data and rule number on MPRM performance when using MolRuleLoss**. **a,** We expect that MolRuleLoss would confer relatively small gains in molecular property prediction accuracy upon expanding the dimensionality of the training data. **b,** The data splitting and SSR extraction process used for studying whether MolRuleLoss improves MPRM accuracy at large training dataset size using the melting point (MP) dataset. The MP dataset contains 275K entries (with each entry comprising a SMILES structure and a temperature value). We trained GEM models with and without MolRuleLoss under different data volumes (0.5K, 5K, 50K, and 100K). **c,** Accuracy comparisons based on RMSE and $R^2$. **d,** Transferring rules from a large dataset with MolRuleLoss in theory improves accuracy of MPRM, which is analogous to the "transfer learning" concept. **e,** RMSE comparison for the "no rules", MP0.5K-rules, and MP100K-rules test iterations of the GEM model using the MP dataset.

## Including low standard deviation rules with MolRuleLoss boosts the accuracy and generalizability of MPRMs for "activity cliff" and OOD molecules

Recall the two overfitting scenarios for DL-based MPRMs mentioned in the introduction: 1) struggling with ID "activity cliffs" (AC) molecules[37, 39] and 2) struggling with OOD molecules[36, 40, 53]. OOD problems can be further divided into structure-OOD and property-OOD (Fig. 4a). Given MolRuleLoss's demonstrated capacity to boost MPRM prediction accuracy, we next examined whether MolRuleLoss improves the generalizability for MPRMs when predicting AC and OOD molecules. We used a lipophilicity prediction task to assess the AC problem. Biologically, consider that an apparently simple substructure substitutions can lead to dramatic changes in lipophilicity (Fig. 4b), thus potentially preventing a compound from passing a membrane. To construct an AC test set, we selected molecule pairs that have a similarity (Morgan fingerprint) higher than 0.75 and a lipophilicity change greater than 1.0 ($\Delta$logP=1.0, an order of magnitude difference). We then put one molecule from each pair into the training set and the other into the test set; the final test set comprised 290 molecules. The RMSE values of the GEM model with and without MolRuleLoss were 0.627 and 0.674 (Fig. 4c). These results demonstrate that MolRuleLoss improves the generalizability of an MPRM for the AC scenario.

Existing MPRMs frequently exhibit overfitting on OOD test sets for properties like permeability and binding affinity[53, 54] . Building on MolRuleLoss's demonstrated performance boost based on extracted SSRs, we investigated whether the SSRs learned and selected for ID training data are generalizable for OOD prediction tasks. We used the aforementioned MP prediction task, specifically using molecules with an MP lower than 231 °C (representing the top 10% highest MP molecules) for training and using molecules with an MP greater than 231 °C for the test set (Fig. 4d). To minimize the influence of SSR errors, we specifically investigated whether incorporating SSRs with low STD values ($STD_{max}$=0.27) via MolRuleLoss enhances MPRM accuracy. The RMSE values for the GEM model with and without MolRuleLoss on the OOD test set were 98.959 and 93.189 (Fig. 4e). Note that although the model with MolRuleLoss did outperform the model without it, the performances of both were catastrophically bad for the OOD molecules. These results support that low-$STD_{max}$ SSRs derived from a narrow training dataset enable an MPRM to "extrapolate" to a broader data scope.

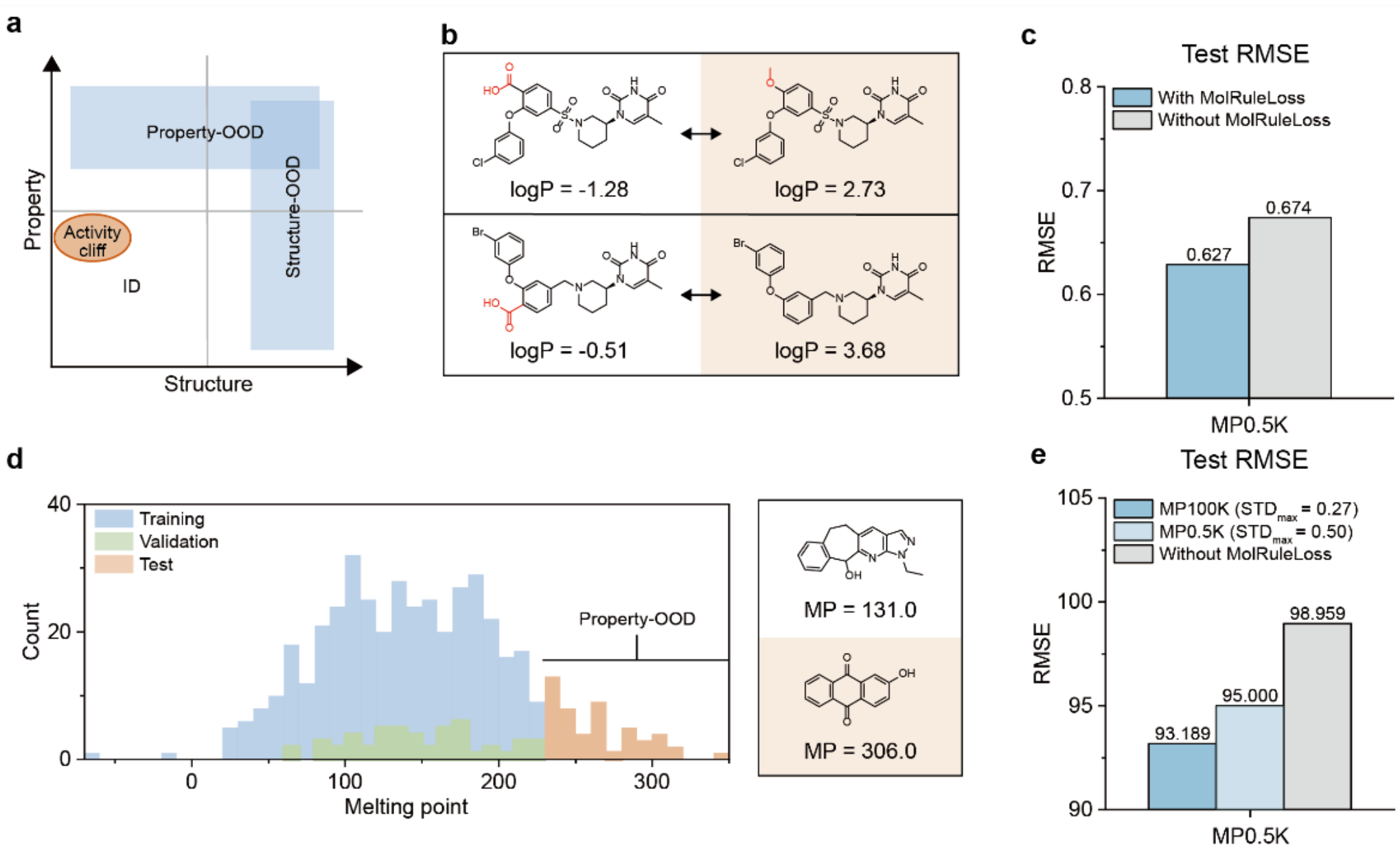

**Figure 4 | Low standard deviation rules improve accuracy and generalizability of MPRMs for activity cliff and out-of-distribution molecules.** **a,** Definition of "Activity cliff" (AC) and out-of-distribution (OOD) molecules. We used two axes to describe the regions for AC and OOD molecules. AC molecules are ID molecules with large property variations stemming from apparently small structural changes. Two categories of OOD molecules are considered: i. Exhibiting large structural difference to the molecules in the training dataset ("structure-OOD"), *e.g.*, molecules have no common substructures to molecules in the training dataset; ii. property values out of the range of the training dataset ("property-OOD"). **b,** Examples for the AC problem using the lipophilicity prediction task. When we use scaffold split, we already considered structure-OOD problem. To construct an AC test set, we selected molecule pairs that have a similarity (Morgan fingerprint) higher than 0.75, yet have a lipophilicity change greater than 1.0 ($\Delta$logP=1.0, an order of magnitude difference). We then put one molecule from each pair into the training set and the other into the test set. **c,** RMSE comparison for a GEM model with or without MolRuleLoss, using the AC test set of the lipophilicity prediction task. **d,** Examples for the OOD problem using the MP prediction task. **e,** RMSE comparison for the GEM model with high-quality rules ($STD_{max}$=0.27), low-quality rules ($STD_{max}$=0.50) or without MolRuleLoss on the OOD test set.

As our low-$STD_{max}$ rule set is neither complete nor deterministic ($STD_{max}$=0), we sought to design a task in which the rules are complete and deterministic, to quantitatively explore potential error reduction. We here employed a molecular weight (MW) prediction, which is calculated deterministically from a molecule's SMILES string by summing atomic masses, making it an inherently error-free molecular property. MW prediction is distinct from the aforementioned lipophilicity, solubility and MP properties, which are determined using experimental measurements, and thus inevitably contain errors. We constructed an MW dataset comprising 2,700 molecules (2,160 for training and validation) by sampling 5 molecules per integer MW from 160 to 700 Dalton[55] (an MW range for drug-like molecules) from the ChEMBL dataset, with entries ranging from 160–600 Dalton as the training dataset, but ranging from 600 Dalton to 700 for the test set (*i.e.*, an OOD test set).

The RMSE value of the GEM model without MolRuleLoss was 29.507, which is catastrophically high. The 66 SSRs derived from our MW dataset represents the difference between the atomic masses of elements (delta is an exact value, $STD_{max}$ = 0.000), which are deterministic. We found that the RMSE of the GEM model with MolRuleLoss dramatically decreased, from 29.507 to 0.007, and the $R^2$ improved from -0.050 to 1.000 (Fig. 5a,b). We also trained GEM models with and without MolRuleLoss at different data ratios, showing that much less data are required when using MolRuleLoss than without it (Fig. 5c). Additionally, when artificial noise is added to the atomic mass, which increases $STD_{max}$, we observe that the RMSE values for the model with MolRuleLoss increase nearly linearly with the noise, suggesting that $STD_{max}$ defines the upper performance limit of our model (Fig. 5d). These results based on GEM models support that the challenges AI models currently face in handling OOD data can be addressed by acquiring SSRs from ID data and leveraging them for extrapolation to OOD contexts.

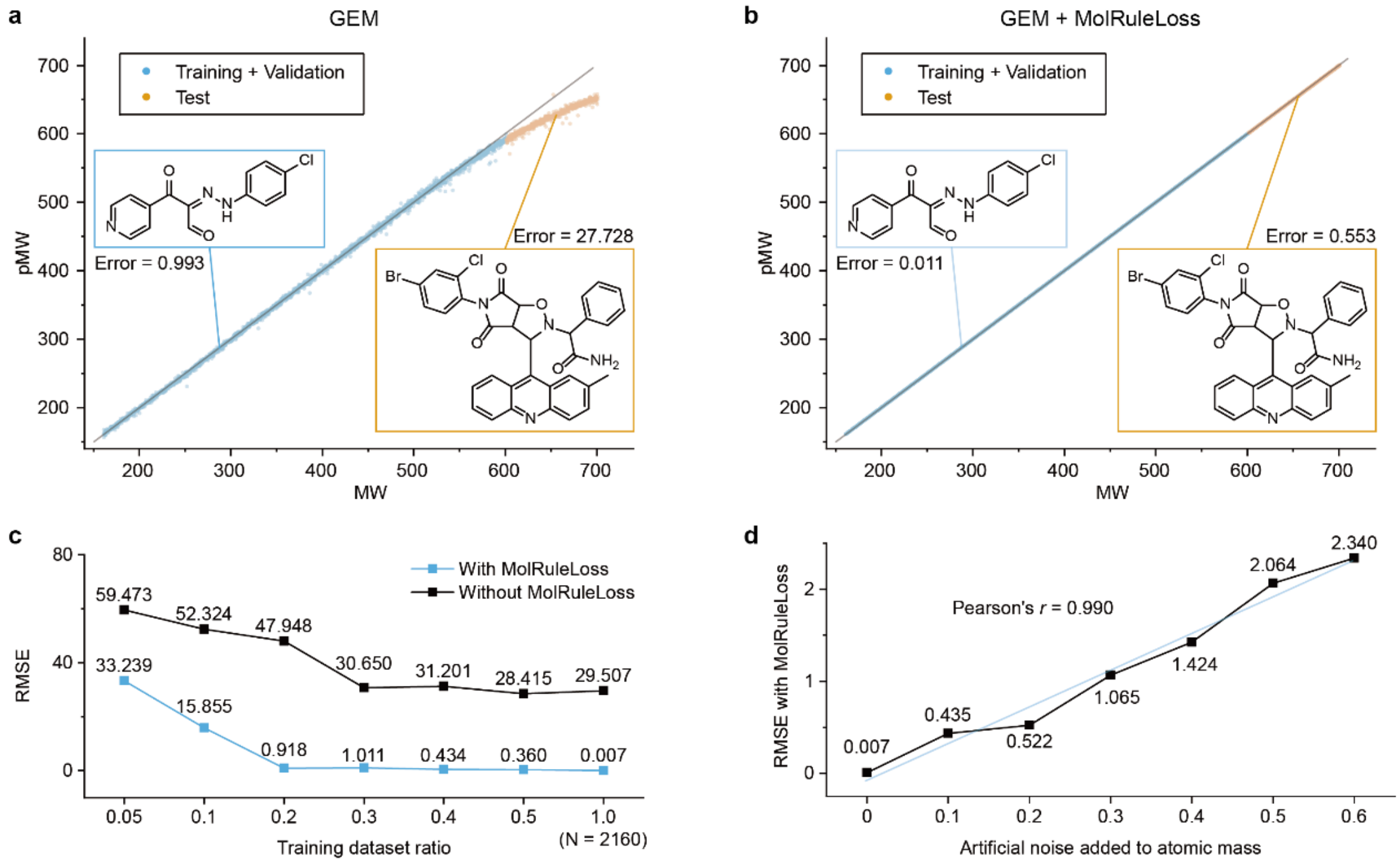

**Figure 5 | Including deterministic rules with MolRuleLoss boosts the accuracy of MPRMs for the molecular weight prediction task**. **a-b,** Molecular weight (MW) OOD prediction using the GEM model with and without MolRuleLoss. MW is calculated deterministically from a molecule's SMILES string by summing atomic masses, making it an inherently error-free property. We constructed a MW dataset comprising 2,700 molecules by sampling 5 molecules per integer MW from 160 to 700 Dalton from the ChEMBL dataset, with entries ranging from 160-600 Dalton as the training and validation dataset (in blue) but ranging from 600 Dalton to 700 Dalton for the test set (in orange). The RMSE value of the GEM model without MolRuleLoss is 29.507, which is catastrophically high; while the RMSE of the GEM model with MolRuleLoss decreased dramatically, from 29.507 to 0.007. **c,** Performance of GEM models with and without MolRuleLoss at different data ratios, **d,** Performance of the GEM model with MolRuleLoss under different levels of artificial noise added to the atomic mass.

## The relationship between the $STD_{max}$ of SSRs and the error of an MPRM

We have shown that incorporating low $STD_{max}$ (high-quality) SSRs via MolRuleLoss enhances the accuracy and generalizability of MPRMs. Specifically, our MP prediction task supports that MPRM with low $STD_{max}$ SSRs are more accurate than MPRM with high $STD_{max}$ SSRs; and our MW prediction task shows that at an extreme condition of $STD_{max}$=0.000, the error of the MPRM is close to 0. The large gain for the OOD with $STD_{max}$=0.000 prompted us to revisit the $STD_{max}$ ranges for earlier prediction tasks, and we noted a trend: model error increases when $STD_{max}$ values increase. We therefore propose the conjecture that the $STD_{max}$ values of the SSRs are positively correlated with the MPRM's error. A mathematical proof supporting this conjecture is provided in Text S2.

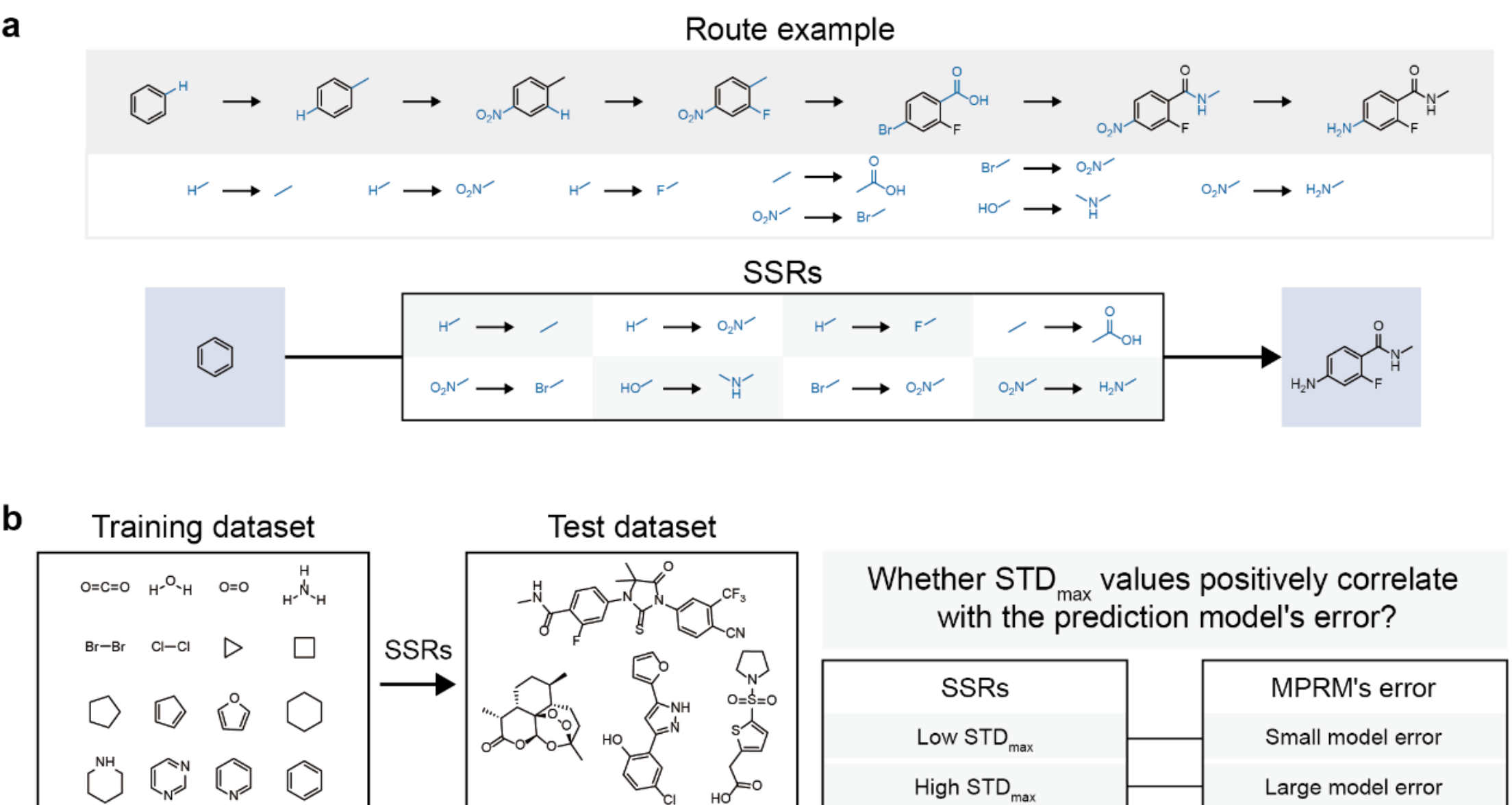

**Figure 6 | A conjecture regarding the relationship between the $STD_{max}$ value of SSRs and MPRM error. a,** A given test set can be obtained by applying a series of SSRs to the training dataset. **b,** When SSRs have a large $STD_{max}$ value, predicting a given property using an MPRM is difficult. If this property is "predictable" ( that is, when the model error is smaller than a threshold value), then $STD_{max}$ is positively correlated with this model error. We offer a mathematical proof for this conjecture (SI), providing a theoretical foundation for our MolRuleLoss framework.

In theory, a test set of synthesizable organic compounds could be obtained via applying SSRs to the molecules from the training dataset (Fig. 6). Our conjecture establishes a theoretical foundation for the MolRuleLoss framework, formally justifying the relevance of SSRs and offering guidance for future efforts to leverage such rules to improve the performance of MPRMs. Briefly, this trend emphasizes the positive utility of adding SSRs under bounded-error conditions; moreover, there is a direct, consistent link between rule variation ($STD_{max}$) and model performance, offering a strong justification for using SSRs to improve model prediction accuracy. Specifically, a high $STD_{max}$ value for SSRs of a given property indicates that an MPRM prediction for this property would have large errors[56] (*e.g.*, as observed with binding affinity).

**Discussion**

In this study, we developed MolRuleLoss, a substructure-substitution-rule-informed framework that implements partial derivative constraints of SSRs derived from the MMPA method. MolRuleLoss consistently boosts the accuracy of multiple MPRMs across diverse molecular property regression tasks. It bears emphasis that even modest accuracy gains will manifest as large improvements for downstream applications, particularly in addressing data scarcity in AI-aided drug discovery. We evaluated the accuracy of MPRMs with or without MolRuleLoss across diverse molecular property regression tasks on the MoleculeNet benchmark dataset, and found that it improved the performance of SOTA MPRMs like GEM and UniMol.

In a melting point prediction task, we show that both the dimensionality and quality of rules each contribute to the magnitude of prediction accuracy gains obtained upon adding MolRuleLoss to an MPRM. Our results support that MolRuleLoss improves the generalizability of an MPRM for "activity cliff" and OOD molecules. In an OOD evaluation for MW prediction, we observed a dramatic reduction in RMSE, suggesting that MPRMs with MolRuleLoss can achieve accurate predictions when deterministic rules are included. We propose a conjecture that the $STD_{max}$ values of the SSRs are positively correlated with the MPRM's error, for which we provide a mathematical proof. Ultimately, MolRuleLoss offers a framework that can be applied as a bolt-on to diverse MPRMs, supporting a wide range of applications including data scarcity scenarios in ADMET optimizations.

In principle, the MolRuleLoss approach for boosting prediction performance should be applicable for other molecular properties, such as permeability and binding affinity. While advances such as AlphaFold3[57] now enable prediction of protein-ligand complex structures, drug discovery efforts still require reliable ranking of candidate small molecules by binding affinity[58-60]. Data augmentation strategies, *e.g.*, increasing data volume and adding synthetic data, have been shown to offer improvements in accuracy for MPRMs[61, 62]. However, data augmentation strategies do not necessarily improve the generalizability involving property-OOD molecules[36, 39-41]. Rule constraints seem likely to improve the generalizability of MPRMs for for property predictions of OOD molecules in permeability and binding affinity tasks. Beyond GEM and UniMol, we anticipate that MolRuleLoss will likely improve generalizability when added onto other MPRMs such as VisNet and attentive FP[21, 63].

Our work highlights several directions for enhancing the accuracy of MPRMs through rule-based partial derivative constraints. We have so far focused on SSRs derived from the MMPA method, and the number of rules is limited when the training dataset size is small. Future extensions could incorporate alternative rule sources. Notably, while we have currently focused solely on pairwise substructure substitution, the MolRuleLoss framework should accommodate other rule formats, for example those involving substitution of more than one substructure. Beyond the domain of molecular property prediction, MolRuleLoss may be applied to other regression tasks (e.g., oral bioavailability). Finally, MolRuleLoss may be appliable to molecular entities including proteins or nucleic acids. To achieve this expanded application scope, one could define substructures of proteins and use AI models to learn potential SSRs from a supervised model.

## Methods

### Dataset

*MoleculeNet*

We sourced molecular datasets from MoleculeNet t[50], focusing on three benchmark tasks: Lipophilicity (logP, n=4,200), ESOL solubility (logS, n=1,150), and FreeSolv hydration free energies (ΔG_solv, n=650). Molecular structures were standardized using RDKit (v2022.09) [64] through a multi-step pipeline that included charge neutralization, salt removal, and aromaticity normalization to Kekulé form. To rigorously evaluate out-of-distribution (OOD) generalization, we implemented two complementary splitting strategies. For scaffold-based OOD evaluation, molecules were clustered by Bemis-Murcko scaffolds using Butina clustering with a Tanimoto similarity cutoff of 0.7 based on Morgan fingerprints (radius 2), with the 10% least populated clusters reserved for testing to maximize structural novelty. For property-based OOD evaluation, molecules were sorted by target property values, with the top and bottom 10% extremes forming the test set to simulate prediction challenges for property outliers. To specifically test extreme extrapolation capabilities, we constructed a dedicated molecular weight prediction dataset from ChEMBL v30[65], spanning 160–700 Da with uniform sampling, where the test set exclusively contained molecules >600 Da (n=300) and inputs were restricted to counts of 12 common atoms (H, C, N, O, F, P, S, Cl, Br, I, B, Si).

*Melting Point Dataset*

Our melting point (MP) dataset was extracted from USPTO patents (1976-2014) using Tetko et al.'s text-mining pipeline[66]. Suspicious records were excluded, including MPs >500 °C, ranges >50 °C, or inverted ranges, removing 1,498 entries from grants and 426 from applications. Duplicate entries (ΔMP ≤ 1 °C) were consolidated, eliminating 366,532 records. For compounds with multiple valid measurements, the value closest to the median was retained to preserve data integrity and patent traceability. Molecules failing descriptor calculation were excluded. Experimental accuracy was estimated at σ = 35 °C based on 18,058 duplicate measurements, accounting for polymorphism and extraction variances. For the drug-like space (50–250 °C, covering >90% of compounds), error refined to σ = 32 °C. The final curated dataset provides a diverse MP collection for predictive modeling. Overlap with established benchmarks was removed to ensure test set independence.

**Butina Clustering Implementation**

We applied the Butina clustering algorithm to identify structurally homogeneous groups within our molecular dataset[67]. This method partitions compounds based on pairwise structural similarity, ensuring that all molecules within a cluster exhibit a minimum similarity to the cluster centroid. Compounds in SMILES format were converted into binary RDKit RDK5 fingerprint vectors, capturing key substructural features. Pairwise structural similarities were computed using the Tanimoto coefficient:

$$Tc = \frac{A \cap B}{A \cup B}$$

This generated a symmetric similarity matrix, subsequently converted to a distance matrix via D = 1 − Tc.

Molecules were ranked by their number of neighbors (compounds within a similarity threshold). The molecule with the most unassigned neighbors was selected as a centroid, and all its neighbors were assigned to the new cluster. These members were then excluded from future centroid selection or cluster membership. Remaining unassigned molecules were classified as singletons. This algorithm efficiently produces clusters with high internal structural consistency.

**Model Architectures**

To enhance the generalization capability of molecular property prediction, we integrated Physics-Informed Neural Networks (PINNs) [68, 69] into the molecular representation learning framework. PINNs represent a class of neural networks that incorporate prior physical knowledge, typically expressed as partial differential equations (PDEs), ordinary differential equations (ODEs), or other physical constraints, directly into the learning process through a multi-component loss function. This approach reduces the reliance on large volumes of purely experimental data and ensures that predictions adhere to fundamental physical laws. In our implementation, building upon established molecular representation learning models, we designed a feedforward neural network (FNN) architecture to serve as a universal function approximator mapping molecular substructure counts to target properties (e.g., logP, logS, $\Delta G_solv$). The core innovation of PINNs lies in their composite loss function, which simultaneously optimizes both data fidelity and physical consistency.

Two state-of-the-art architectures served as baselines for molecular representation learning. The Geometry-Enhanced Molecular GNN (GEM)[18] processes 2D molecular graphs through an 8-layer Graph Isomorphism Network (GIN) with 32-dimensional hidden states, using atom features that encode atomic number, degree, hybridization, implicit valence, and formal charge. Pretrained via self-supervised bond length and angle prediction tasks, GEM employs a two-layer MLP ($32 \rightarrow 32 \rightarrow 1$) with dropout ($p=0.1$) for property regression. In contrast, UniMol (Universal Molecular Transformer)[20] operates on 3D conformers generated using RDKit's ETKDG method with MMFF94 optimization. This SE(3)-equivariant Transformer architecture uses 512-dimensional embeddings with 8 attention heads and rotary positional embeddings, pretrained on 209 million molecular conformations from QM9 and GEOM datasets. For property prediction, UniMol reduces dimensions through a $512 \rightarrow 128 \rightarrow 1$ MLP with LayerNorm. Both models were implemented in PyTorch with mixed-precision training on NVIDIA A100 GPUs.

**MolRuleLoss Framework**

Our approach integrates chemical domain knowledge through differentiable substructure substitution rules (SSRs), systematically derived from matched molecular pair (MMP) analysis. Using RDKit's MMP implementation (max cut size=13 atoms), we identified all molecular pairs differing by exactly one substructure and computed mean property changes ($\Delta P$) and their standard deviations ($\sigma$) for each substitution pattern. Twenty-nine high-confidence rules were retained ($\sigma < 0.5$ and occurrence $\geq 10$), including common transformations like phenyl → cyclohexyl ($\Delta logP = -0.82 \pm 0.21$) and -OH → $-NH_2$ ($\Delta logS = +0.47 \pm 0.33$). The MolRuleLoss function combines conventional prediction error with a rule consistency term:

$$Loss = Loss_{MSE} + \lambda \times Loss_{SSR}$$

where $Loss_{SSR}$ enforces agreement between model gradients and empirical rules through the term

$$Loss_{SSR} = \sum_{i,j} [(\frac{\partial y}{\partial_{\#} x_i} - \frac{\partial y}{\partial_{\#} x_j}) - \mathrm{rule}(x_i, x_j)]^2$$

Implementation leverages PyTorch's automatic differentiation to compute gradients of the model output y with respect to 29-dimensional substructure count vectors, with these counts concatenated to pretrained representations before final prediction.

Recognizing that the impact of identical substructural substitutions on molecular properties can be significantly modulated by their distinct chemical environments—namely, the broader structural context of the host molecule—we introduce a novel adaptive rule refinement algorithm. This approach addresses the critical limitation of conventional matched molecular pair (MMP) analysis, wherein static transformation rules fail to account for context-dependent effects on property changes.

Our methodology dynamically adjusts rule-based predictions by integrating learned molecular representations with the specific contextual environment of each molecule. The implementation proceeds as follows: For a given substructural transformation rule with an initial predicted property change value $\Delta$, we generate a high-dimensional molecular representation vector h using a pre-trained molecular graph neural network (GNN). This representation comprehensively encodes the structural and electronic features of the entire molecular scaffold surrounding the substitution site.

Simultaneously, we initialize a learnable parameter vector $\theta$ of identical dimension to the molecular representation. The adaptive adjustment term is computed as the dot product between the molecular representation vector and the parameter vector:

$$\delta = h \cdot \theta$$

This term is then added to the base rule value to yield the context-sensitive prediction:

$$\Delta_{Adaptive} = \Delta + \delta$$

During the model training phase, both the molecular representation network (if fine-tuned) and the adaptive parameter vectors are optimized jointly using backpropagation to minimize the difference between predicted and experimental property changes. This learning process enables the model to capture systematic variations in rule applicability across different chemical environments.

Upon convergence, the learned parameter vectors become fixed and are utilized in all subsequent predictions. This approach effectively transforms static, context-agnostic rules into dynamic, adaptive rules that respond to the specific chemical environment of each molecular transformation, thereby significantly enhancing prediction accuracy for complex molecular systems.

## Training and Evaluation

All models were trained using Adam optimization (lr=1e-4, $\beta_1$=0.9, $\beta_2$=0.999) with decoupled weight decay (1e-5) and cosine annealing with warm restarts (15 epochs per cycle). Training employed batch size 32 with dropout (p=0.1), gradient clipping (max norm=5.0), and early stopping (10-epoch patience). Performance was assessed via root mean squared error (RMSE) and coefficient of determination ($R^2$), with five replicates (seeds 1024–1028) to estimate variance. Specialized OOD tests included activity cliff evaluation (290 molecular pairs with Tanimoto similarity >0.75 but ΔlogP >1.0) and molecular weight extrapolation (train: 160–600 Da, test: 600–700 Da). Full reproducibility is ensured through publicly available code, datasets, and configuration files that document all preprocessing steps, model implementations, and hyperparameter choices.

## GEM Model Architecture

The core of our model utilizes a multi-layer GNN architecture based on the message-passing framework. At each layer l, the node representations are updated through the following operations:

Message Passing: For node $v \in V$ at layer l, messages from neighbors $u \in N(v)$ are computed as:

$$m_{u\to v}^{(l)} = \mathrm{MSG}^{(l)}(h_u^{(l-1)}, h_v^{(l-1)}, e_{uv})$$

where $N(v)$ denotes the neighborhood of node $v$, $h_u^{(l-1)}$ and $h_v^{(l-1)}$ are node features from the previous layer, and euvrepresents edge features.

Aggregation: The aggregated message for node $v$ is obtained by:

$$\mathrm{M}_v^{(l)} = \mathrm{AGGREGATE}^{(l)}(\{m_{u\to v}^{(l)} : u \in N(v)\})$$

We employ sum aggregation followed by a multilayer perceptron (MLP), similar to the Graph Isomorphism Network (GIN) approach:

$$a_v^{(l)} = \sum_{u \in N(v)} (h_u^{(l-1)} + h_v^{(l-1)} + e_{uv})$$

Update: The node representation is updated through a COMBINE function:

$$h_v^{(l)} = \mathrm{COMBINE}^{(l)}(h_v^{(l-1)}, M_v^{(l)}) = \mathrm{MLP}^{(l)}(a_v^{(l)})$$

where $\mathrm{MLP}^{(l)}$ denotes a 2-layer multilayer perceptron with hidden size of 32.

The model consists of 8 GNN layers with hidden dimension 32, providing sufficient depth for capturing complex molecular patterns. Layer Normalization applied before each message passing step. Graph Size Normalization to address variable graph sizes. Residual Connections to facilitate gradient flow in deep architectures

For graph-level prediction tasks, we generate a global graph representation through a readout function that aggregates all node representations from the final GNN layer:

$$h_G = \mathrm{READOUT}(\{h_v^{(L)} : v \in V\})$$

We employ global mean pooling as the readout function, resulting in a 32-dimensional graph embedding.

To enhance predictive performance, we integrate domain-specific molecular substructure features with the learned graph representation. Let fsub∈Rmdenote the m-dimensional vector encoding the counts of various molecular substructures. The final integrated representation is obtained by concatenation:

$$h_{final} = \text{Concat}(h_G, f_{substructure})$$

This combined representation is then processed through a prediction head consisting of a 3-layer MLP with hidden size of 128:

$$\hat{y} = \text{MLP}_{prediction}(h_{final})$$

where the output dimension depends on the specific prediction task. This architecture allows the model to simultaneously leverage both learned graph embeddings and expert-defined molecular features for enhanced predictive performance.

**Matched Molecular Pair Analysis (MMPA) Implementation**

We employed the mmpdb package (v3.1.3) for Matched Molecular Pair Analysis[70] to systematically extract structural transformation rules and associated property change profiles from our molecular dataset. The workflow consisted of four phases:

Compounds in SMILES format were processed using the mmpdb fragmentation algorithm, which cuts 1-3 non-ring bonds to decompose structures into a constant core fragment and variable R-group fragments. This captures all possible matched molecular pairs (MMPs). Fragmented representations were indexed to identify compound pairs sharing identical constant cores but differing in one variable fragment. This generated a database of structural transformations, with each transformation cataloged alongside its experimental property measurements. Each transformation was characterized by its rule environment using two representations: 1) SMARTS patterns (RDKit circular fingerprints, radius 2) for machine-readable environment definitions; 2) Pseudo-SMILES strings for human-interpretable representations

This enables differentiation of transformations across distinct chemical contexts. For each transformation-environment combination, we computed mean property change, standard deviation, standard error, and fraction of pairs exceeding property change thresholds. Transformations with ≥10 matched pairs and well-defined distributions (low standard error) were retained as predictive rules.

## Data Availability

The datasets of molecular properties affected by this work can be found in the public website MoleculeNet at https://moleculenet.org/. MW dataset can be found in the Github repository at https://github.com/fanxiaoyu0/Uni-Mol.

## Code Availability

The code used for this paper can be found in the Github repository at https://github.com/fanxiaoyu0/Uni-Mol.

## Acknowledgments

This work was supported by the National Natural Science Foundation of China (No.22473067), Beijing Frontier Research Center for Biological Structure (No. 041500002), Tsinghua University Initiative Scientific Research Program (No.20231080030), and the Tsinghua-Peking University Center for Life Sciences (No.20111770319).

## Competing interests

The authors declare no competing interests.

## Author Information

Correspondence and requests for materials should be addressed to B.T. (boxuetian@mail.tsinghua.edu.cn).

## Author Contributions

X.F. implemented the methods, developed the codes and performed relevant experiments. L.G.prepared all the figures. R. J. provided literature review and contributed to polish the paper. Y.T. contributed to the conception and design of the work. Z. Y. contributed to the data analysis. B.T. designed the project, analyzed data, wrote and revised the manuscript. All authors edited and revised the manuscript. All authors wrote the paper and read and approved the final paper.

## References

[1] Niu, Z., Xiao, X., Wu, W., et al. PharmaBench: enhancing ADMET benchmarks with large language models. *Scientific Data*, **11**, 985. (2024).

[2] Swanson, K., Walther, P., Leitz, J., et al. ADMET-AI: a machine learning ADMET platform for evaluation of large-scale chemical libraries. *Bioinformatics*, **40**, btae416. (2024).

[3] Yi, J., Shi, S., Fu, L., et al. OptADMET: a web-based tool for substructure modifications to improve ADMET properties of lead compounds. *Nature Protocols*, **19**, 1105-1121. (2024).

[4] Fu, L., Shi, S., Yi, J., et al. ADMETlab 3.0: an updated comprehensive online ADMET prediction platform enhanced with broader coverage, improved performance, API functionality and decision support. *Nucleic Acids Res*, **52**, W422-W431. (2024).

[5] Gu, Y., Yu, Z., Wang, Y., et al. admetSAR3. 0: a comprehensive platform for exploration, prediction and optimization of chemical ADMET properties. *Nucleic Acids Res*, **52**, W432-W438. (2024).

[6] Walters, W. P.& Barzilay, R. Applications of deep learning in molecule generation and molecular property prediction. *Acc Chem Res*, **54**, 263-270. (2021).

[7] Zeng, X., Xiang, H., Yu, L., et al. Accurate prediction of molecular properties and drug targets using a self-supervised image representation learning framework. *Nat Mach Intell*, **4**, 1004-1016. (2022).

[8] Zheng, Y., Koh, H. Y., Ju, J., et al. Large language models for scientific discovery in molecular property prediction. *Nat Mach Intell*, **7**, 437-447. (2025).

[9] Shimakawa, H., Kumada, A.& Sato, M. Extrapolative prediction of small-data molecular property using quantum mechanics-assisted machine learning. *npj Comput Mater*, **10**, 11. (2024).

[10] Nuñez-Andrade, E., Vidal-Daza, I., Gomez-Bombarelli, R., et al. Embedded Morgan Fingerprints for more efficient molecular property predictions with machine learning. Preprint at https://doi.org/10.26434/chemrxiv-2025-6hfp8 (2025).

[11] Gallegos, M., Vassilev-Galindo, V., Poltavsky, I., et al. Explainable chemical artificial intelligence from accurate machine learning of real-space chemical descriptors. *Nat Commun*, **15**, 4345. (2024).

[12] Chuang, K. V., Gunsalus, L. M.& Keiser, M. J. Learning molecular representations for medicinal chemistry: miniperspective. *J Med Chem*, **63**, 8705-8722. (2020).

[13] Rasool, A., Ul Rahman, J.& Uwitije, R. Enhancing molecular property prediction with quantized GNN models. *J Cheminform*, **17**, 81. (2025).

[14] Wu, Z., Wang, J., Du, H., et al. Chemistry-intuitive explanation of graph neural networks for molecular property prediction with substructure masking. *Nat Commun*, **14**, 2585. (2023).

[15] Chen, D., Gao, K., Nguyen, D. D., et al. Algebraic graph-assisted bidirectional transformers for molecular property prediction. *Nat Commun*, **12**, 3521. (2021).

[16] Qiao, J., Jin, J., Wang, D., et al. A self-conformation-aware pre-training framework for molecular property prediction with substructure interpretability. *Nat Commun*, **16**, 4382. (2025).

[17] Wang, S., Guo, Y., Wang, Y., et al. SMILES-BERT: large scale unsupervised pre-training for molecular property prediction. *In Proc 10th ACM Conference on Bioinformatics, Computational Biology, and Health Informatics* (ACM-BCB ’19,2019).

[18] Fang, X., Liu, L., Lei, J., et al. Geometry-enhanced molecular representation learning for property prediction. *Nat Mach Intell*, **4**, 127-134. (2022).

[19] Yang, Z., Wang, L., Huang, T., et al. Q-GEM: quantum chemistry knowledge fusion geometry-enhanced molecular representation for property prediction. *Advanced Science*, 1-11. (2025).

[20] Zhou, G., Gao, Z., Ding, Q., et al. Uni-mol: a universal 3d molecular representation learning framework. *In Proc 11th International Conference on Learning Representations* (OpenReview.net,2023).

[21] Wang, Y., Wang, T., Li, S., et al. Enhancing geometric representations for molecules with equivariant vector-scalar interactive message passing. *Nat Commun*, **15**, 313. (2024).

[22] Liu, L., He, D., Fang, X., et al. GEM-2: next generation molecular property prediction network with many-body and full-range interaction modeling. Preprint at https://doi.org/10.48550/arXiv.2208.05863 (2022).

[23] Ji, X., Wang, Z., Gao, Z., et al. Exploring molecular pretraining model at scale. *In 38th Conference on Neural Information Processing Systems* (NeurIPS,2024).

[24] Dou, B., Zhu, Z., Merkurjev, E., et al. Machine learning methods for small data challenges in molecular science. *Chem Rev*, **123**, 8736-8780. (2023).

[25] Yin, T., Panapitiya, G., Coda, E. D., et al. Evaluating uncertainty-based active learning for accelerating the generalization of molecular property prediction. *J Cheminform*, **15**, 105. (2023).

[26] Ji, Y., Zhang, L., Wu, J., et al. DrugOOD: out-of-distribution dataset curator and benchmark for AI-aided drug discovery–a focus on affinity prediction roblems with noise annotations. *In Proc 37th AAAI Conference on Artificial Intelligence* **37**, 8023-8031 (AAAI,2023).

[27] Kim, J., Willette, J., Andreis, B., et al. Robust molecular property prediction via densifying scarce labeled data. *In Proc 42th International Conference on Machine Learning* (PMLR,2025).

[28] Shen, W. X., Cui, C., Su, X., et al. Activity Cliff-Informed contrastive learning for molecular property prediction. Preprint at https://doi.org/10.21203/rs.3.rs-2988283/v2 (2024).

[29] Dablander, M. Investigating graph neural networks and classical feature-extraction techniques in activity-cliff and molecular property prediction. Preprint at https://doi.org/10.48550/arXiv.2411.13688 (2024).

[30] Zhang, Y., Li, X., Peng, N., et al. CROSS: a cross-dimensional representation optimization self-supervised learning framework for activity cliffs prediction. *In Proc 31st International Conference on Neural Information Processing* **2286**, 31-48 (ICONIP,2024).

[31] Wan, Y., Wu, J., Hou, T., et al. Multi-channel learning for integrating structural hierarchies into context-dependent molecular representation. *Nat Commun*, **16**, 413. (2025).

[32] Jiang, J., Zhang, R., Yuan, Y., et al. NoiseMol: a noise-robusted data augmentation via perturbing noise for molecular property prediction. *J Mol Graph Model*, **121**, 108454. (2023).

[33] Wang, Z., Jiang, T., Wang, J., et al. Molecular connectivity index-based data augmentation for molecular property prediction. *IEEE Transactions on Computational Biology and Bioinformatics*. (2025).

[34] Cheng, Z., Xiang, H., Ma, P., et al. MaskMol: knowledge-guided molecular image pre-

training framework for activity cliffs with pixel masking. Preprint at https://doi.org/10.48550/arXiv.2409.12926 (2024).

[35] Kuang, T., Liu, P.& Ren, Z. Impact of domain knowledge and multi-modality on intelligent molecular property prediction: a systematic survey. *Big Data Mining and Analytics*, **7**, 858-888. (2024).

[36] Antoniuk, E. R., Zaman, S., Ben-Nun, T., et al. BOOM: benchmarking out-of-distribution molecular property predictions of machine learning models. Preprint at https://doi.org/10.48550/arXiv.2505.01912 (2025).

[37] Van Tilborg, D., Alenicheva, A.& Grisoni, F. Exposing the limitations of molecular machine learning with activity cliffs. *J Chem Inf Model* **62**, 5938-5951. (2022).

[38] Wang, Y., Xiong, J., Xiao, F., et al. LogD7. 4 prediction enhanced by transferring knowledge from chromatographic retention time, microscopic pKa and logP. *J Cheminf*, **15**, 76. (2023).

[39] Tamura, S., Miyao, T.& Bajorath, J. Large-scale prediction of activity cliffs using machine and deep learning methods of increasing complexity. *J Cheminform*, **15**, 4. (2023).

[40] Deng, J., Yang, Z., Wang, H., et al. A systematic study of key elements underlying molecular property prediction. *Nat Commun*, **14**, 6395. (2023).

[41] Pang, C., Tong, H. H.& Wei, L. Advanced deep learning methods for molecular property prediction. *Quantitative Biology*, **11**, 395-404. (2023).

[42] Fang, Y., Zhang, Q., Zhang, N., et al. Knowledge graph-enhanced molecular contrastive learning with functional prompt. *Nat Mach Intell*, **5**, 542-553. (2023).

[43] Terven, J., Cordova-Esparza, D.-M., Romero-González, J.-A., et al. A comprehensive survey of loss functions and metrics in deep learning. *Artificial Intelligence Review*, **58**, 195. (2025).

[44] Heid, E., Greenman, K. P., Chung, Y., et al. Chemprop: a machine learning package for chemical property prediction. *J Chem Inf Model*, **64**, 9-17. (2024).

[45] Qin, Y., Li, C., Shi, X., et al. MLP-based regression prediction model for compound bioactivity. *Front Bioeng Biotechnol*, **10**, 946329. (2022).

[46] Wang, S., Sankaran, S., Wang, H., et al. An expert's guide to training physics-informed neural networks. Preprint at https://doi.org/10.48550/arXiv.2308.08468 (2023).

[47] Krishnapriyan, A., Gholami, A., Zhe, S., et al. Characterizing possible failure modes in physics-informed neural networks. *In Proc 35th Conference on Neural Information Processing Systems* **34**, 26548-26560 (NeurIPS,2021).

[48] Yang, Z., Shi, S., Fu, L., et al. Matched molecular pair analysis in drug discovery: methods and recent applications. *J Med Chem*, **66**, 4361-4377. (2023).

[49] Yang, K., Swanson, K., Jin, W., et al. Analyzing learned molecular representations for property prediction. *J Chem Inf Model* **59**, 3370-3388. (2019).

[50] Wu, Z., Ramsundar, B., Feinberg, E. N., et al. MoleculeNet: a benchmark for molecular machine learning. *Chem Sci*, **9**, 513-530. (2018).

[51] Sushko, I., Novotarskyi, S., Körner, R., et al. Online chemical modeling environment (OCHEM): web platform for data storage, model development and publishing of chemical information. *J Comput Aided Mol Des*, **25**, 533-554. (2011).

[52] Weiss, K., Khoshgoftaar, T. M.& Wang, D. A survey of transfer learning. *Journal of Big Data*, **3**, 9. (2016).

[53] Fooladi, H., Vu, T. N. L.& Kirchmair, J. Evaluating machine learning models for molecular property prediction: performance and robustness on out-of-distribution data. Preprint at https://doi.org/10.26434/chemrxiv-2025-g1vjf-v2 (2025).

[54] Shi, Y., Xu, W.& Hu, P. Out of distribution learning in bioinformatics: advancements and challenges. *Briefings in Bioinformatics*, **26**, bbaf294. (2025).

[55] Bickerton, G. R., Paolini, G. V., Besnard, J., et al. Quantifying the chemical beauty of drugs. *Nat Chem*, **4**, 90-98. (2012).

[56] Zhang, H., Liu, X., Cheng, W., et al. Prediction of drug-target binding affinity based on deep learning models. *Comput Biol Med*, **174**, 108435. (2024).

[57] Abramson, J., Adler, J., Dunger, J., et al. Accurate structure prediction of biomolecular interactions with AlphaFold 3. *Nature*, **630**, 493-500. (2024).

[58] Chen, Z., Peng, B., Zhai, T., et al. Generating 3D binding molecules using shape-conditioned diffusion models with guidance. *Nat Comput Sci*, **4**, 899-909. (2024).

[59] An, L., Said, M., Tran, L., et al. Binding and sensing diverse small molecules using shape-complementary pseudocycles. *Science*, **385**, 276-282. (2024).

[60] Yu, J., Li, Z., Chen, G., et al. Computing the relative binding affinity of ligands based on a pairwise binding comparison network. *Nat Comput Sci*, **3**, 860-872. (2023).

[61] bin Javaid, M., Gervens, T., Mitsos, A., et al. Exploring data augmentation: Multi-task methods for molecular property prediction. *Comput Chem Eng*, **201**, 109253. (2025).

[62] Jiang, J., Li, Y., Zhang, R., et al. INTransformer: Data augmentation-based contrastive learning by injecting noise into transformer for molecular property prediction. *J Mol Graph Model*, **128**, 108703. (2024).

[63] Xiong, Z., Wang, D., Liu, X., et al. Pushing the boundaries of molecular representation for drug discovery with the graph attention mechanism. *J Med Chem*, **63**, 8749-8760. (2019).

[64] Landrum, G. Rdkit documentation. *Release*, **1**, 4. (2013).

[65] Gaulton, A., Bellis, L. J., Bento, A. P., et al. ChEMBL: a large-scale bioactivity database for drug discovery. *Nucleic Acids Res*, **40**, D1100-D1107. (2012).

[66] Tetko, I. V., M. Lowe, D.& Williams, A. J. The development of models to predict melting and pyrolysis point data associated with several hundred thousand compounds mined from PATENTS. *J Cheminf*, **8**, 2. (2016).

[67] Butina, D. Unsupervised data base clustering based on daylight's fingerprint and Tanimoto similarity: A fast and automated way to cluster small and large data sets. *J Chem Inf Model*, **39**, 747-750. (1999).

[68] Wang, S., Sankaran, S., Wang, H., et al. An expert's guide to training physics-informed neural networks. Preprint at https://doi.org/10.48550/arXiv.2308.08468 (2023).

[69] Raissi, M., Perdikaris, P.& Karniadakis, G. E. Physics-informed neural networks: A deep learning framework for solving forward and inverse problems involving nonlinear partial differential equations. *J Comput Phys*, **378**, 686-707. (2019).

[70] Dossetter, A. G., Griffen, E. J.& Leach, A. G. Matched molecular pair analysis in drug discovery. *Drug Discov Today*, **18**, 724-731. (2013).

# Supplementary Information

## Improving the accuracy and generalizability of molecular property regression models with a substructure-substitution-rule-informed framework

Xiaoyu Fan[1], Lin Guo[1], Ruizhen Jia[1], Yang Tian[1], Zhihao Yang[1], Weihao Li[2] and Boxue Tian[1#]

*[1]MOE Key Laboratory of Bioinformatics, State Key Laboratory of Molecular Oncology, Beijing Frontier Research Center for Biological Structure, School of Pharmaceutical Sciences, Tsinghua University, Beijing, 100084, China. [2]Department of Statistics and Data Science, Tsinghua University, Beijing, 100084, China.*

[#] To whom correspondence should be addressed:

Boxue Tian: boxuetian@mail.tsinghua.edu.cn

# Text S1. Proof of "Tian Conjecture" on the relationship between the upper bound of deviation values of substructure substitution rules and the prediction error of a molecular property regression model

## Overview

Assume any two compounds are interconvertible via a sequence of rules, the conjecture states that while a molecular property regression model (MPRM) has a test set error smaller than $e$ (meaning that the model can predict the property of any compound in the test set with sufficient accuracy), then, there exists a a set of substructure substitution rules (SSRs), R, and a generated set S⊆S(T,R), where T is the training dataset and S is a test set, such that: For every rule r in R, the maximum deviation values for the property changes, $\sigma_r$ satisfies

$$\sigma_r \leq 3e$$

This justifies that the upper bound of standard deviation values property changes upon using substructure substitution rules affect the model's error. Our conjecture may apply for deep learning-based molecular property prediction of organic molecules that related to drug discovery.

## 1. Definitions

Let the following mathematical objects be defined:

***Chemical space***

Let C be the set of synthesizable chemical compounds that defined by a set of chemical reactions, and each reaction can be viewed as a substructure substitution. Each compound c∈C is represented as a molecular graph.

***Training dataset***

A finite subset T⊂C, for example, |T|=10,000.

***Test set***

A finite subset S⊂C, for example, |S|=1,000,000.

***Property Functions***

True property: ***P***: C→R

Model prediction: ***F***: C→R

***Model Accuracy (Assumption 1)***

For a given error threshold ***e***>0, for compound c in the test set S⊂C:

$$| P(c) - F(c) | \leq e \tag{1}$$

***Substructure substitution rules***

A **rule** r=(s,s′) replaces substructure s with s′ in a compound c.

The **applicable context** for r is $C_r$ = {c∈C|c contains s}.

A **rule set** R is a collection of such rules.

**Generated Set**: The set G(T,R)⊆C consists of all compounds reachable from T via finite applications of rules in R.

**Property Change and Standard Deviation**:

For a rule r and context c∈Cr, the property change is:

$$\Delta \boldsymbol{P}(r,c) = \boldsymbol{P}(r(c)) - \boldsymbol{P}(c)$$

The **deviation** of an SSR Δ**P** over applicable contexts is:

$$\sigma_r = |\Delta P(r,c)|$$

***Model Stability (Assumption 2)***

For a given error threshold e > 0, for compound c in the test set S⊂C, for a rule r, the prediction change is small, *i.e.*,

$$|F(r(c)) - F(c)| \leq e \tag{2}$$

## *2. Proof Outline*

The small generalization error e (Assumption 1) implies that the model F approximates P well on S. The small perturbation error e (Assumption 2) implies that the model F is stable enough. These two assumptions force the variability of property changes ΔP to be small when rules are applied in S.

We relate ΔP and ΔF and then use triangle inequality:

$$\sigma_r = |P(r(c)) - P(c)| = |P(r(c)) - F(r(c)) + F(r(c)) - F(c) + F(c) - P(c)|$$

$$\leq |P(r(c)) - F(r(c))| + |F(r(c)) - F(c)| + |F(c) - P(c)|$$

$$\leq e + e + e$$

$$= 3e$$

## Text S2. Investigations on the out-of-distribution problems of various deep learning models

### *Overview*

We define two out-of-distribution (OOD) scenarios: Structure-OOD comprising distribution shifts at the molecular structure level, and Property-OOD, which reflect poorly overlapping label distributions.

### *Predictive accuracy dramatically decreases on Structure-OOD and Property-OOD data*

To illustrate the poor generalizability of current deep learning models for predicting properties of OOD molecules, we first compared the structural distributions of molecules across different datasets. Specifically, we examined the Lipophilicity (Lipo), ESOL, Freesolv, and BACE datasets from MoleculeNet, a benchmark collection for molecular property prediction, and compared them to corresponding datasets in Chemdiv, a widely used molecular library for in silico drug screening. Using Morgan fingerprints for molecular representation and t-SNE for dimensionality reduction (Fig. S1a), we observed differences in the structural distributions between the MoleculeNet and Chemdiv datasets, indicating Structure-OOD conditions.

To identify which molecules and properties in the Chemdiv data fell outside the MoleculeNet distribution, we trained multiple deep learning models with different random seeds on the Lipo dataset. We then used Morgan fingerprints to calculate pairwise similarity between molecules. Clustering was performed with the Butina algorithm from the RdKit package, sorted in descending order by cluster size. We selected the tail 20% of molecules as the test set to simulate real-world drug library screening scenarios where a varying proportion of molecules exhibit low structural similarity to the training set. The GEM model was chosen as the molecular feature encoder. We found that the root-mean-square error (RMSE) between predicted and actual lipophilicity was 0.514 for the validation set and 0.843 for the test set. In comparison, directly reporting the average lipophilicity resulted in an RMSE of 1.2. These findings suggest that using the GEM model to predict lipophilicity in a drug library is likely to yield erroneous predictions for a significant proportion of molecules.

We then investigated the issue of Property-OOD, using the molecular weight prediction task as an example. Property-OOD refers to the distribution offset at the label level. For instance, the molecular weight distributions of the MoleculeNet datasets compared to the Chemdiv library are shown in Fig. S1b. There is a significant difference between the molecular weight distributions of the ESOL and Freesolv datasets and the Chemdiv library, indicating Property-OOD.

To assess the impact of label-level distribution shift on prediction performance, we conducted the following analyses. Molecules (150k+) were sourced from the Tsinghua University School of Pharmacy library[https://cpt.tsinghua.edu.cn], through uniform sampling of 20 molecules per molecular weight integer interval. Samples with molecular weights greater than 400 Dalton were allocated to the test set, while the remaining samples (up to 500 Dalton) were randomly divided into training or validation sets at a 3:1 ratio. The GEM model was selected as the deep learning model for its relatively high accuracy and extremely high training speed. The prediction results are shown in Fig. S2a. The validation set had an RMSE of 4.86, while the RMSE of the test set was 39.13. These results highlight the poor performance of deep learning models on Property-OOD data, underscoring their limited generalizability.

***Poor generalizability on Property-OOD data is universal across model architectures and tasks***

To elucidate the reasons behind the poor predictive accuracy of the GEM model in OOD scenarios, we investigated four small molecule property prediction models with distinctly different architectures: XGBoost, a machine learning model that uses molecular descriptors as input; SMILES-BERT, a language-based model that employs SMILES as input; GEM, a graph neural network that utilizes molecular topology along with bond length and angle information; and Uni-Mol, a Transformer model that uses molecular 3D structures as input. It is important to note that SMILES-BERT, GEM, and Uni-Mol are pre-trained on large-scale unlabeled molecule datasets. Additionally, GEM and Uni-Mol were state-of-the-art models on the MoleculeNet dataset at the time of their publication, making these four models a reasonably representative sample set in terms of model structure and predictive ability. We performed molecular weight prediction on the same test and validation sets used to identify Property-OOD errors. All models exhibited generalization errors to varying extents.

We further tested each model on various regression tasks to assess generalization errors across different molecular property prediction challenges. The tasks included relative molecular mass prediction (representing intrinsic properties of molecules), lipophilicity prediction (representing physicochemical properties), compound-protein affinity prediction (representing biological activity), and QM7 prediction (representing quantum chemical properties). The results, summarized in Fig. S2b, show that poor generalizability on OOD data is common across different tasks and model architectures. These findings indicate that the issue of poor generalizability extends beyond a single model or task, highlighting a shared challenge in molecular property prediction when dealing with OOD data.

***Increasing* dataset *size has limited effect on OOD data***

As poor predictive performance may be attributed to insufficient training data (citations), we next tested whether increasing the training set scale improves model generalizability. As increasing the amount of experimentally supported training data for various molecular properties can be time- and labor-intensive, whereas molecular weight can be directly calculated, we repeated the above MW prediction task after training the models with a larger sample set. That is, we increased sample size from 1 molecule per integer of MW to 300 molecules per integer MW, expanding the dataset from 540 molecules to 162000 molecules. However, we found that simply training with more data could not completely solve this problem (Fig. S2c). While increasing sample size from 1 to 20 molecules per integer molecular weight resulted in a significantly decreased RMSE from 53.8 to 6.59, increasing sample size from 20 to 300 molecules per integer molecular weight resulted in only a modest decline in RMSE, from 6.59 to 6.27. These results suggested that increasing the size of training data does not fully resolve poor generalizability on Property-OOD data.

### ***Double-layer MLP has poor generalizability in MW prediction***

To verify this approach to addressing the problem of OOD-associated errors, while reducing or eliminating possible interference from other factors, we selected the simplest task (MW prediction), the simplest model (double-layer MLP) and the simplest feature (atom count of each element) for our experiments. To this end, we constructed a MW dataset by sampling 5 molecules per integer MW from the CHEMBL database. This sampling scheme yielded a dataset with MW range from [160 Dalton, 700 Dalton], comprising 2700 molecules. Those molecules with MW >600 Dalton were classified as the test set, while the remaining samples were randomly divided into training and validation sets in a 3:1 ratio, as above.

Tests of a double-layer MLP showed that had an RMSE of 9.933 on the test set, indicating that double-layer MLP did not show the strong generalizability. Moreover, adjusting hyperparameters such as learning rate, weight decay, batch normalization, number of neurons, activation function, random seeds, etc. did not resolve this poor accuracy in predictions of OOD samples. By contrast, the resulting linear regression had an error close to zero on the test set (RMSE = 0.002), which suggested strong generalizability.

### ***Model with partial* derivative *constraints***

Given the above poor performance of double-layer MLP, we next investigated whether and how multi-layer MLPs could be modified to yield a generalizable model. Analysis of molecules in the test set with the highest prediction errors by the double-layer MLP showed that these molecules commonly contained chemical elements that were relatively rare in the training set, such as B, I, P and Si (Fig. S3a). We therefore speculated that the relative scarcity of such atoms in the training set resulted in incomplete learning of their respective atomic masses as contributions the MW prediction for OOD molecules. To resolve this issue, we hypothesized that inductive bias based on our well-established prior knowledge of these elements could be injected into multi-layer deep learning models to improve generalizability.

Seeking to enhance model generalizability, we examined the relationship among data volume, model size (model parameters), and the extent of human-induced bias. In scenarios with limited data, introducing human knowledge may benefit predictive accuracy, while requiring fewer model parameters. Conversely, as the volume of available data increases, the benefit of human-induced bias diminishes, and models with stronger inductive bias are more constrained and show lower predictive accuracy in such contexts. Therefore, to improve outcomes on our small, original dataset, we considered possible strategies for incorporating inductive bias that accounted for 1) types of human knowledge we had available that would benefit prediction accuracy, and 2) how to inject such prior knowledge to the model.

For this purpose, we selected molecular weight, which is relative the atomic mass individual atoms comprising a molecule. Abstracting these principles into mathematical language, using Si as an example in calculating MW, we found that they can be applied as actually constraints on partial derivatives, as follows:

$$\frac{\partial \text{ MoleculeWeight}}{\partial \, \#_{Si}} = 28.086$$

That is, the partial derivative of molecular weight, with respect to the number of silicon atoms, should be equal to 28. To apply this inductive bias that requires constraining partial derivatives, we add the constraint to the loss function by modifying the loss function of PINN (Fig. S3b). Incorporating the above partial derivative restrictions for the four rarest atoms (B, I, P, Si) into a double-layer MLP, we observed that RMSE decreased from 9.93 to 0.355 in the test set (Fig. S3c). These results indicated that MW prediction accuracy was obviously improved.

In addition, we also observed an interesting phenomenon. Heatmap visualization of parameters in the double-layer MLP with or without partial derivative restrictions (Fig. S3d) indicated that the incorporation of PINN resulted in more regular distribution of weight across parameters, while the parameters appear disorganized in the model lacking partial derivative constraint. This may mean that models with stronger generalization ability have simpler weight structures; of course, this is only intuitive speculation.

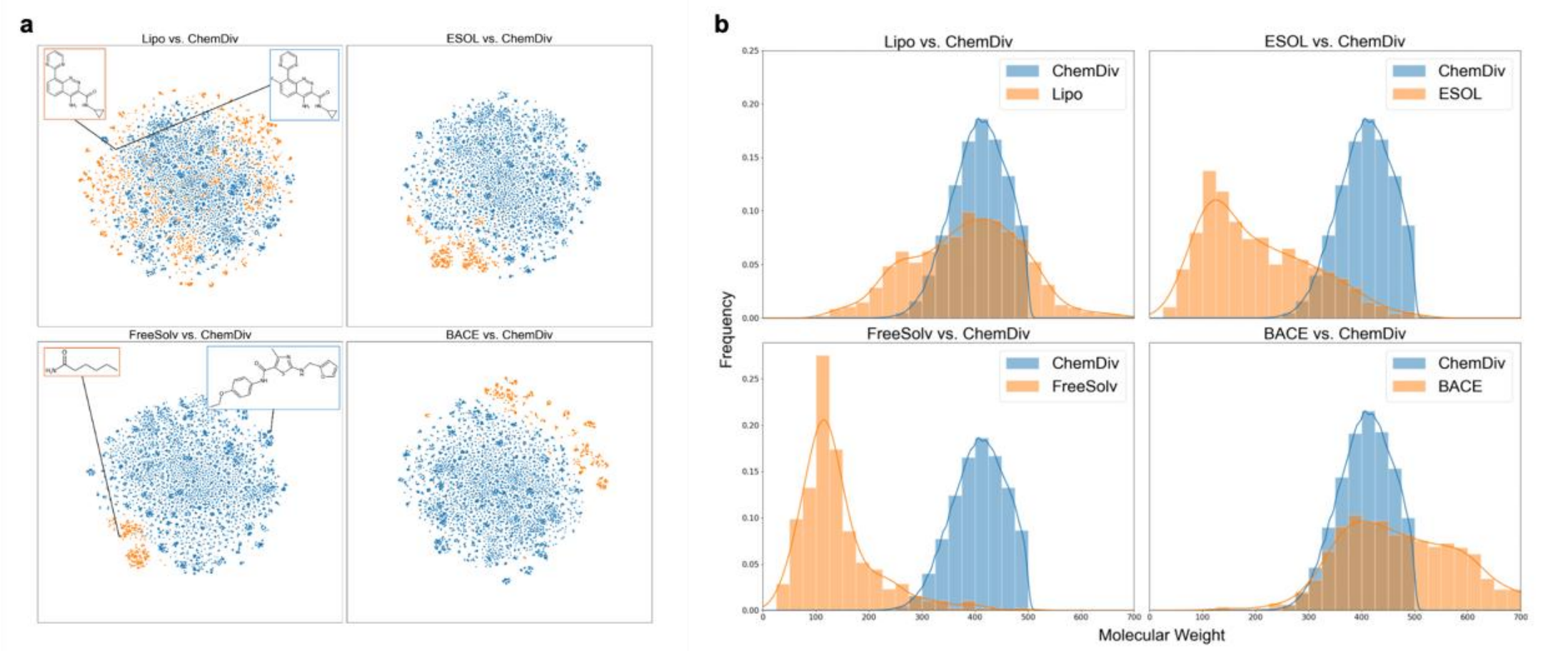

**Figure S1. Comparison of molecular structure and weight distributions between MoleculeNet and Chemdiv datasets.**

**a**, T-SNE plots of molecular structures represented by Morgan fingerprints for four datasets from MoleculeNet (Lipophilicity, ESOL, Freesolv, and BACE) compared to the Chemdiv dataset. The distinct distributions of molecules in the ESOL, Freesolv, and BACE datasets (orange) relative to Chemdiv molecules (blue) indicate poor overlap and Structure-OOD characteristics. **b**, Histograms of the molecular weight distributions in four MoleculeNet datasets (Lipophilicity, ESOL, Freesolv, and BACE) superimposed on the molecular weight distribution in the Chemdiv dataset. The distributions for the ESOL and Freesolv datasets (orange) exhibit significant deviation from the Chemdiv distribution (blue), highlighting discrepancies at the label level and illustrating Property-OOD characteristics.

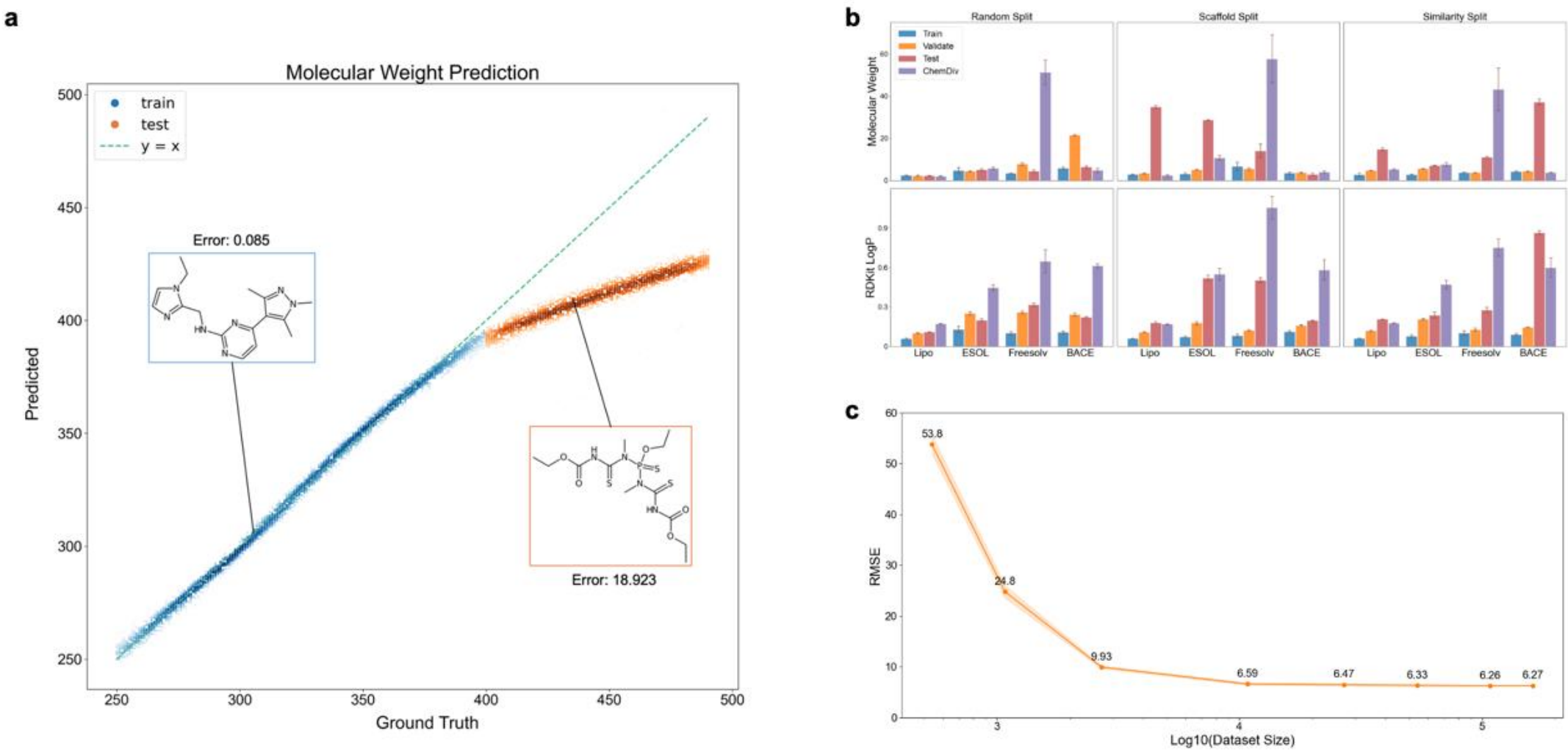

**Figure S2. Impact of label distribution shift, model generalization performance, and training set size on MW prediction.**

**a**, Impact of label level distribution shift on performance in the MW prediction task. Scatter plot of predicted versus calculated MWs. A GEM model trained on a dataset comprised of molecules <400 Dalton (blue) was tested on a dataset containing molecules ≥400 Dalton (orange). The dashed green line at y=x represents complete agreement between predicted and true MW values. Predictions for OOD samples (*i.e.*, ≥400 Dalton) in the test set show obvious deviation from the true values, indicating poor generalizability and highlighting the challenges of Property-OOD in MW prediction. **b**, Generalization performance of various deep learning models across MW and lipophilicity prediction tasks. Bar graphs of root mean square error (RMSE) in MW and lipophilicity predictions using the Random, Scaffold, and Similarity data splitting methods. XGBoost, SMILES-BERT, GEM, and Uni-Mol model performance was assessed on training (blue), validation (red), test (orange), and Chemdiv (green) datasets. Notably, all models exhibited greater RMSE on the Chemdiv dataset, highlighting limited generalizability on OOD data. **c**, Effect of training set size on generalizability in MW prediction. Plot of RMSE values in the MW prediction task with increasing training set size from 1 to 300 molecules per integer MW. RMSE decreases significantly from 53.8 to 6.59 as the sample size increases from 1 to 20 molecules per interval, but further increase in sample size, up to 300 molecules per interval, results in limited reduction of RMSE to 6.27, indicating that larger training sets do not enable complete generalizability.

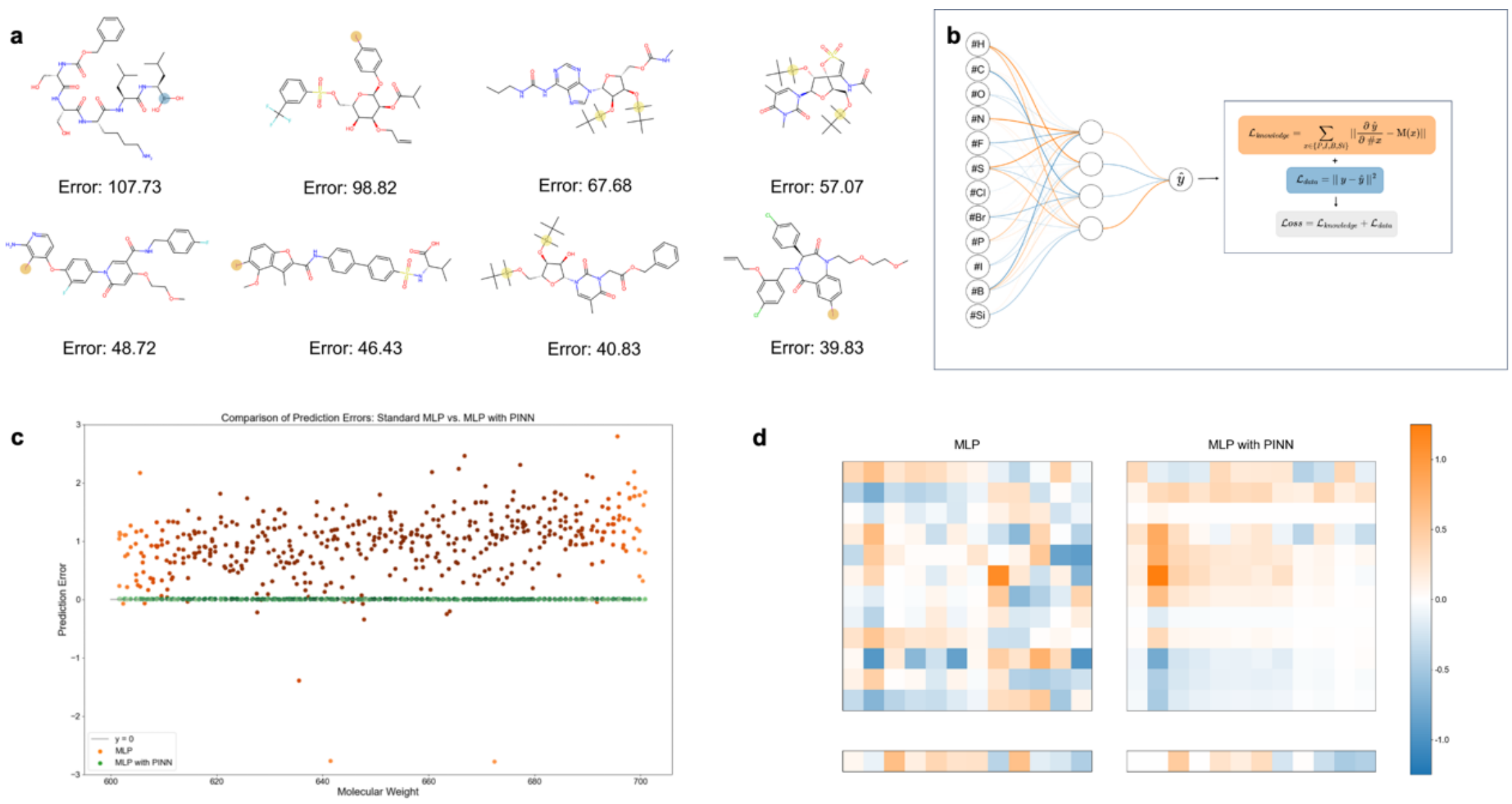

**Figure S3. Model performance and the impact of partial derivative constraints on MW prediction.**

**a**, Examples of molecules with the highest prediction errors by the double-layer MLP model. Each molecule's error (the absolute difference between predicted and true MW) is shown below its structure. Notably, all of these molecules contain elements that were relatively rare in the training set, such as B, I, P, and Si, suggesting that the model struggled to accurately learn the contributions of these elements to MW. **b**, Incorporation of partial derivative constraints into the loss function for improved model generalizability. Schematic illustrating the modification of the loss function to include partial derivative constraints based on domain knowledge. The constraints are applied to elements like Si for MW, ensuring the model learns the correct contributions of these features. The combined loss function integrates both prediction error and gradient constraints to enhance model generalizability. **c**, Comparison of prediction errors for MW between standard MLP and MLP with PINN. Scatter plot showing the prediction errors for MW using a standard double-layer MLP (brown) and a double-layer MLP with partial derivative constraints (PINN, green). The x-axis represents the true MW, and the y-axis represents the prediction error. Incorporating partial derivative constraints significantly reduces the prediction errors, demonstrating improved generalizability for the MLP with PINN. **d**, Heatmap visualization of parameters in the double-layer MLP with and without partial derivative constraints. Heatmaps showing the distribution of parameter weights in a standard double-layer MLP (left) and a double-layer MLP with partial derivative constraints (PINN, right). The incorporation of PINN results in a more regular and organized distribution of weights across parameters, while the parameters appear more disorganized in the model lacking these constraints.